\documentclass[preprint,12pt]{elsarticle}
\usepackage{amssymb,amsmath,amsthm}
\usepackage[mathscr]{euscript}
\usepackage[T1]{fontenc}
\usepackage{cleveref}
\usepackage{booktabs}
\usepackage{url}
\usepackage[linesnumbered,ruled,commentsnumbered]{algorithm2e}
\usepackage{csvsimple}
\usepackage{pgfplots, pgfplotstable}
\usepackage{tikz}
\usepackage{multicol,multirow}
\usepackage{tabularx}
\usepackage{enumitem}
\usepackage{comment}
\usepackage{rotating}
\usetikzlibrary{math}
\usepgfplotslibrary{groupplots, fillbetween}
\usetikzlibrary{positioning,chains}
\usetikzlibrary{arrows}
\usetikzlibrary{decorations}
\usetikzlibrary{decorations.markings}
\usetikzlibrary{shapes}
\usetikzlibrary{intersections}
\usetikzlibrary{patterns}
\usetikzlibrary{pgfplots.statistics}
\pgfplotsset{compat=newest}
\usetikzlibrary{calc, backgrounds}

\theoremstyle{definition}  
\newtheorem{definition}{Definition}
\theoremstyle{remark}   
\newtheorem{remark}{Remark}
\theoremstyle{plain}

\newtheorem{example}{Example}
\newtheorem{problem}{Open Problem}

\DeclareMathOperator*{\argmin}{arg\,min}

\newcommand{\directgolib}{\texttt{DIRECTGOLib v2.0}}

\newcommand{\direct}{\texttt{DIRECT}}
\newcommand{\directl}{\texttt{DIRECT-l}}
\newcommand{\plor}{\texttt{PLOR}}
\newcommand{\hddirect}{\texttt{HD-DIRECT}}
\newcommand{\abcd}{\texttt{ABCD}}

\newcommand{\idtcgl}{\texttt{I-DTC-GL}}
\newcommand{\xdtcgl}{\texttt{X-DTC-GL}}
\newcommand{\nmso}{\texttt{NMSO}}

\newcommand{\directt}{\texttt{tDIRECT}}
\newcommand{\directsqp}{\texttt{DIRECT-SQP}}

\newcommand{\dirmin}{\texttt{DIRMIN}}
\newcommand{\birmin}{\texttt{BIRMIN}}
\newcommand{\glccluster}{\texttt{glcCluster}}
\newcommand{\directrev}{\texttt{DIRECT-rev}}
\newcommand{\directc}{\texttt{I-DTC-IOl}}
\newcommand{\plorc}{\texttt{I-DTC-RPl}}

\let\oldnl\nl
\newcommand{\nonl}{\renewcommand{\nl}{\let\nl\oldnl}}
\newcommand{\algrule}[1][.2pt]{\par\vskip.5\baselineskip\hrule height #1\par\vskip.5\baselineskip}

\newcommand{\PreserveBackslash}[1]{\let\temp=\\#1\let\\=\temp}
\newcolumntype{C}[1]{>{\PreserveBackslash\centering}p{#1}}
\newcolumntype{R}[1]{>{\PreserveBackslash\raggedleft}p{#1}}
\newcolumntype{L}[1]{>{\PreserveBackslash\raggedright}p{#1}}

\emergencystretch=\maxdimen
\tikzmath
{
	function symlog(\x,\a){
		\yLarge = ((\x>\a) - (\x<-\a)) * (ln(max(abs(\x/\a),1)) + 1);
		\ySmall = (\x >= -\a) * (\x <= \a) * \x / \a ;
		return \yLarge + \ySmall ;
	};
	function symexp(\y,\a){
		\xLarge = ((\y>1) - (\y<-1)) * \a * exp(abs(\y) - 1) ;
		\xSmall = (\y>=-1) * (\y<=1) * \a * \y ;
		return \xLarge + \xSmall ;
	};
}

\makeatletter
\tikzset{
	nomorepostactions/.code={\let\tikz@postactions=\pgfutil@empty},
	mymark/.style 2 args={decoration={markings,
			mark= between positions 0 and 1 step (1/11)*\pgfdecoratedpathlength with{%
				\tikzset{#2,every mark}\tikz@options
				\pgfuseplotmark{#1}%
			},
		},
		postaction={decorate},
		/pgfplots/legend image post style={
			mark=#1,mark options={#2},every path/.append style={nomorepostactions}
		},
	},
}
\makeatother

\tikzset{
	s_a1/.style = {mymark={o}{draw=blue,solid,mark size=2pt},blue,line width=0.75pt,opacity=0.8},
	s_a2/.style = {mymark={*}{draw=red,solid,mark size=2pt},red,line width=0.75pt,opacity=0.8},
	s_a3/.style = {mymark={triangle}{draw=green,solid,mark size=2pt},green,line width=0.75pt,opacity=0.8},
	s_a4/.style = {mymark={triangle*}{draw=orange,solid,mark size=2pt},orange,line width=0.75pt,opacity=0.8},
	s_a5/.style = {mymark={square}{draw=purple,solid,mark size=2pt},purple,line width=0.75pt,opacity=0.8},
	s_a6/.style = {mymark={square*}{draw=brown,solid,mark size=2pt},brown,line width=0.75pt,opacity=0.8},
	s_a7/.style = {mymark={diamond}{draw=cyan,solid,mark size=2pt},cyan,line width=0.75pt,opacity=0.8},
	s_a8/.style = {mymark={diamond*}{draw=magenta,solid,mark size=2pt},magenta,line width=0.75pt,opacity=0.8},
	s_a9/.style = {mymark={pentagon}{draw=yellow,solid,mark size=2pt},yellow,line width=0.75pt,opacity=0.8},
	s_a10/.style = {black, line width=0.75pt, very thick},
	s_a11/.style = {mymark={+}{draw=gray,solid,mark size=2pt},gray,line width=0.75pt,opacity=0.8},
}

\definecolor{color1}{HTML}{960018}
\definecolor{color2}{HTML}{FF6F61}
\definecolor{color3}{HTML}{6B486B}
\definecolor{color4}{HTML}{FFBF00}
\definecolor{color5}{HTML}{0070FF}
\definecolor{color6}{HTML}{FDEE00}
\definecolor{color7}{HTML}{000000}
\definecolor{cc1}{HTML}{000000}
\definecolor{cc2}{HTML}{000000}
\definecolor{cc3}{HTML}{000000}
\definecolor{cc4}{HTML}{000000}
\definecolor{cc5}{HTML}{000000}
\definecolor{cc6}{HTML}{000000}
\definecolor{cc7}{HTML}{FFFFFF}
\pgfplotsset{
	selective show sum on top/.style={/pgfplots/scatter/@post marker code/.append code={
			\ifnum\coordindex=#1
			\node[at={(normalized axis cs: \pgfkeysvalueof{/data point/x}, \pgfkeysvalueof{/data point/y})}, anchor=west, text=black, opacity=1, font=\scriptsize,] {\pgfmathprintnumber{\pgfkeysvalueof{/data point/x}}};
			\fi
		},
	}, selective show sum on top/.default=0
}

\begin{document}

\begin{frontmatter}

\title{A practical \direct-type algorithm for medium-scale black-box global optimization}

\author[1]{Linas Stripinis} \ead{linas.stripinis@mif.vu.lt}
\author[1]{Remigijus Paulavi\v{c}ius\corref{cor1}} \ead{remigijus.paulavicius@mif.vu.lt}

\address[1]{Vilnius University Institute of Data Science and Digital Technologies, Akademijos 4, Vilnius, Lithuania, LT-08663}
\cortext[cor1]{Corresponding author}

\begin{abstract}
The \direct{} algorithm is a deterministic global optimization method known for its versatility and balanced exploration-exploitation strategy. 
However, \direct-type algorithms are primarily effective for low-dimensional problems and often exhibit slow convergence as dimensionality increases, limiting their applicability to more complex optimization tasks.
To address this limitation, this paper introduces \xdtcgl, a novel \direct-type algorithm that incorporates dynamic partitioning and hybridization techniques.
The dynamic partitioning approach adaptively refines the search space based on local one-dimensional surrogate models, enabling rapid subdivision of promising hyper-rectangles. 
The hybridization strategy selectively employs a hill-climbing method to exploit promising regions identified by the surrogate models. 
Extensive experiments on four diverse benchmark suites demonstrate that \xdtcgl{} significantly outperforms existing \direct-type baselines, achieving improvements of ${\sim}12\%$ in solvability and ${\sim}27\%$ in solution quality. 
Performance-profile analyses indicate the fastest convergence on up to ${\sim}40\%$ of instances, the best runtime performance on ${\sim}17\%$ of problems, and competitive overall execution times.
By improving performance within the partition-based framework, these advances strengthen the algorithm’s competitiveness in state-of-the-art black-box optimization.
\end{abstract}


\end{frontmatter}

\section{Introduction}

This paper discusses a single-objective optimization problem with box constraints:
\begin{equation}\label{eq_OptProblem}
	\begin{aligned}
		& \min_{\mathbf{x}\in \mathcal{X}} && f(\mathbf{x}).
	\end{aligned}
\end{equation}
The problem is to minimize the objective function $f(\mathbf{x})$, where $\mathbf{x} \in \mathbb{R}^n$ and $f:\mathbb{R}^n \rightarrow \mathbb{R}$, subject to the constraints that $\mathbf{x}$ lies within a hyper-cube:
\begin{equation}\label{eq_OptDomain}
	\mathcal{X} = \{ \mathbf{x} \in \mathbb{R}^n: l_j \leq x_j \leq u_j, j = 1, \dots, n\}.
\end{equation}
Here, the vectors $\mathbf{l} \in \mathbb{R}^n$ and $\mathbf{u} \in \mathbb{R}^n$ represent the lower and upper bounds of the search space for the optimization variable $\mathbf{x}$.
The $f$ is Lipschitz continuous and can be non-linear, non-convex, multi-modal, and non-differentiable.
The Lipschitz constant of $f$ is unknown.
We assume that $f$ can only be evaluated at any given point in the feasible region $\mathcal{X}$.
In other words, we cannot access additional information about the objective function, such as its gradient or Hessian.
This is known as Black-Box Optimization (BBO), where the structure of $f$ is unknown, unexploitable, or nonexistent.

Black-box optimization problems, where the objective function is evaluated through simulations, experiments, or complex computations, arise in diverse fields such as engineering design (e.g., airfoil shape optimization), machine learning (e.g., hyperparameter tuning), finance (e.g., portfolio optimization), and drug discovery (e.g., molecular structure identification).
Due to the diverse nature of these applications, there is a growing need for efficient and robust BBO algorithms that can handle various problem characteristics.

Numerous techniques are available to solve BBO problems. 
A recent classification~\cite{Stork2022_taxonomy} categorizes these algorithms into six classes: \textit{exact}, \textit{hill-climbing}, \textit{trajectory}, \textit{population}, \textit{surrogate}, and \textit{hybrid}. 
For each class, we cite well-known, canonical, and representative algorithms to illustrate the underlying optimization principles rather than provide an exhaustive list of current state-of-the-art methods.
\begin{itemize}
	\item \textbf{Exact methods} guarantee finding the global optimum within specified tolerances. Examples include branch-and-bound~\cite{Piyavskii1967, Shubert1972, Paulavicius2009b, Paulavicius2010:ol} and \texttt{DI}viding \texttt{RECT}angles (\direct)~\cite{Jones1993_DIRECT, Jones2021_review, Stripinis2023_book} methods.
	\item \textbf{Hill-climbing methods} focus on greedy exploitation using fast-convergent local optimizers like quasi-Newton methods~~\cite{Fletcher1987_book,JavadEbadi03072023}, Nelder--Mead simplex~\cite{nelder1965simplex}, and proximal bundle methods~\cite{Fadavi2026}.
	\item \textbf{Trajectory algorithms} systematically explore the search space, such as simulated annealing~\cite{Kirkpatrick1983_sa}.
	\item \textbf{Population-based algorithms} operate on a set of potential solutions, including genetic algorithms~\cite{Holland1975} and particle swarm optimization~\cite{kennedy1995particle,FU2023119573}.
	\item \textbf{Surrogate models} replace expensive evaluations with cheaper ones, reducing computational costs, as exemplified by efficient global optimization~\cite{Jones1998, ZENG20221641,BARTZBEIELSTEIN2017154}.
	\item \textbf{Hybrid algorithms} combine components from different classes or use optimization and machine learning to find optimal compositions~\cite{Neri2012memetic, PAN2021304, Kerschke2019_review,DONG2018641}.
\end{itemize}

This paper focuses on \direct-type algorithms, a subclass of exact methods. 
\direct{} is attractive due to its simplicity, ease of implementation, deterministic nature, global convergence guarantees, and minimal hyperparameter tuning~\cite{Jones1993_DIRECT}. 
It effectively balances local and global search (exploitation vs. exploration) by sampling multiple points in each iteration dedicated to both aspects~\cite{Jones2021_review, Stripinis2023_book}. 
A recent review~\cite{Stripinis2024_review} highlights the versatility of \direct-type algorithms in optimizing various applications, including financial portfolios, transportation systems, engineering designs, energy processes, and medical imaging.
Some of the most recent applications we summarized in \Cref{tab:direct_applications}. 
While \direct-type methods are most commonly employed for low- to moderate-dimensional black-box optimization problems, the reported applications exhibit diverse computational requirements. 
For example, TMS electric-field optimization required approximately $2.5{\times}10^3$ function evaluations~\cite{Worbs2025}, whereas car-following model calibration employed a budget of $10^4$ function evaluations~\cite{Li2016_DIRECTsqp}.

\begin{table}
	\small
	\centering
	\caption{Recent applications of DIRECT-type algorithms in diverse domains.}
	\label{tab:direct_applications}
	\begin{tabular*}{\textwidth}{@{\extracolsep{\fill}}lp{0.43\textwidth}p{0.29\textwidth}l}
		\toprule
		\textbf{Ref.} & \textbf{Application} & \textbf{Algorithm(s)} & \textbf{$n$} \\
		\midrule
		$2016$~\cite{Li2016_DIRECTsqp}        & Car-following model calibration        & \texttt{DIRECT-SQP} & $3$--$5$ \\
		$2023$~\cite{Kanayama2023_tDIRECT}    & Atomic cluster structure search        & \texttt{tDIRECT}    & $11$--$71$ \\
		$2023$~\cite{Dapsys_2023}             & Inverse biosensor parameter estimation & \texttt{Aggressive DIRECT} & $3$ \\
		$2025$~\cite{Worbs2025}               & TMS electric-field optimization        & \texttt{DIRECT}, \texttt{DIRECT-l} & $6$--$12$\\
		$2025$~\cite{HVIDSTEN2025108265}      & Biodiesel reaction optimization        & \texttt{DIRECT-l} & $4$ \\
		$2025$~\cite{CHEN2025114514}          & Autonomous-driving model selection     & \texttt{CSD} (stochastic \texttt{DIRECT}) & $2$ \\
		$2025$~\cite{CHEN20251204}            & Blisk tool-orientation optimization    & \texttt{DIRECT} & $2$\\
		\bottomrule
	\end{tabular*}
\end{table}

Recent benchmarking studies~\cite{Stripinis2024_benchmark, Jakub2023} of state-of-the-art BBO methods across multiple algorithmic paradigms, including several top-performing algorithms in recent CEC competitions (e.g., \texttt{EA4eig}~\cite{bujok2022eigen}, \texttt{EBOwithCMAR}~\cite{kumar2017improving}, and \texttt{HSES}~\cite{zhang2018hybrid}), as well as methods ranked among the leading performers like \texttt{L-SHADE} variants~\cite{tanabe2014improving, hadi2021single} indicate that \direct-type algorithms remain highly strong competitors under specific operational settings. 
Their performance is particularly competitive when only moderate evaluation budgets are available (e.g., $5000{\times}n$ function evaluations) and when addressing low-dimensional optimization problems, where they can remain effective even with considerably larger computational budgets. 
As the problem dimensionality increases to moderate levels, their competitive performance becomes more specific to particular problem classes, notably optimization problems that are both non-separable and multi-modal. 
Consequently, establishing a significant performance advancement over the most efficient existing \direct-type baselines implicitly enhances their overall competitiveness against the broader spectrum of state-of-the-art black-box optimizers.

In general, \direct-type algorithms suffer from slow convergence as the dimensionality increases, which restricts their effectiveness on higher-dimensional optimization problems~\cite{Jones2021_review}.\cite{Jones2021_review}.
This slow convergence stems from the need to repeatedly trisect the minimum-containing hyper-rectangle in each dimension to refine the solution, with the convergence rate worsening linearly as dimensionality increases~\cite{Jones2021_review}.  
Additionally, the algorithm explores other hyper-rectangles in each iteration (``global drag''), further increasing the computational cost~\cite{Stripinis2021_pDIRECT_GLce, Tavassoli2015_hddirect}.

Several approaches have been proposed to address these challenges:
\begin{enumerate}
	\item \textbf{Improving the pure \direct-type framework:} Methods like \plor~\cite{Mockus2017_Plor} and \directl~\cite{Gablonsky2001_DIRECTl} prioritize exploitation, but this can delay finding the global minimum~\cite{Jones2021_review, Stripinis2018_DIRECT_GL}.
	\item \textbf{Hybridization:} Combining \direct{} with hill-climbing techniques~\cite{Tao2017_ABC_DIRECT, Jones2001_DIRECT_REV, Kanayama2023_tDIRECT, Liuzzi2010_DIRMIN, Paulavicius2020_BIRMIN, Stripinis2019_DIRECT_GLce} can improve convergence, but excessive hill-climbing can lead to high computational costs~\cite{Kanayama2023_tDIRECT, Liuzzi2010_DIRMIN}.
	\item \textbf{Involving stochastic heuristics:} Techniques like the stochastic ``zoom-in'' (\hddirect)~\cite{Tavassoli2015_hddirect} and block coordinate descent with SQP (\abcd)~\cite{Tao2017_ABC_DIRECT} have been explored, but their effectiveness has been questioned~\cite{Jones2021_review}.
\end{enumerate}

This article introduces \xdtcgl, a novel \direct-type algorithm with two key enhancements to improve convergence as the problem dimensionality increases:

\begin{enumerate}
	\item \textbf{Dynamic partitioning of hyper-rectangles:} This approach adaptively refines the search space based on local one-dimensional surrogate models, enabling rapid subdivision of promising regions.
	\item \textbf{Hybridization with hill-climbing:} This strategy selectively employs a hill-climbing method to exploit promising regions identified by the surrogate models.
\end{enumerate}

\xdtcgl{} constructs local surrogate models within hyper-rectangles to identify promising directions and potential, allowing for swift adjustment of solutions and mitigating the computational cost of global search. The algorithm is tested and validated using four diverse benchmark suites with various problem characteristics.

\subsection{Main contributions and structure of the paper}

This paper makes the following contributions:

\begin{enumerate}
	\item A comprehensive review of techniques for addressing higher-dimensional problems in \direct-type algorithms.
	\item The development of \xdtcgl, a novel and efficient \direct-type algorithm designed to mitigate the curse of dimensionality.
	\item The release of \xdtcgl{} as an open-source resource to ensure reproducibility and reusability of the results.
\end{enumerate}

The paper is structured as follows.
\Cref{sec_review} reviews the original \direct{} algorithm and its extensions for higher-dimensional problems.
\Cref{sec_algorithm} presents the \xdtcgl{} algorithm in detail.
\Cref{sec_experiments} describes the experimental setup and results.
\Cref{sec_conclusions} summarizes the findings and discusses future research directions.

\section{Literature review}
\label{sec_review}

\subsection{Original \direct{} algorithm}

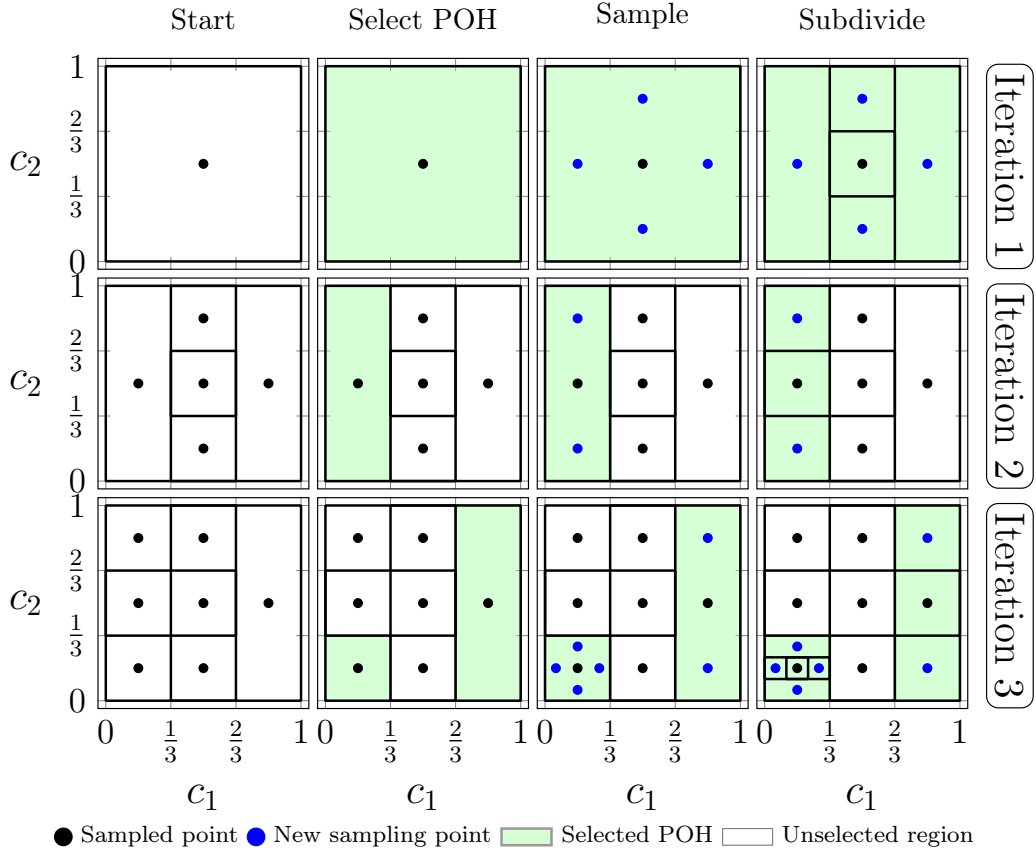
\begin{figure}[ht]
	\centering
	\resizebox{\textwidth}{!}{%
		\begin{tikzpicture}[baseline]
			\begin{groupplot}[group style={group size=4 by 3,x descriptions at=edge bottom, y descriptions at=edge left, horizontal sep=3pt, vertical sep=3pt}, ymin=0, ymax=1, xmin=0, xmax=1, width=0.3\textwidth, height=0.3\textwidth, every axis/.append style={font=\normalsize}, ylabel = {\rotatebox{-90}{$c_2$}}, xlabel = {$c_1$}, label style={font=\large}, enlargelimits=0.04, ytick distance=1/3, xtick distance=1/3, yticklabels={$0$,$0$,$\frac{1}{3}$,$\frac{2}{3}$,$1$}, xticklabels={$0$,$0$,$\frac{1}{3}$,$\frac{2}{3}$,$1$},legend columns=4,legend style={at={(1.1,-0.5)},font=\scriptsize, draw=none, column sep=1pt,}]
				\nextgroupplot[title={\footnotesize Start}]
				\addplot[thick,patch,mesh,draw,black,patch type=rectangle,line width=0.3mm] coordinates {(0,0) (1,0) (1,1) (0,1)};
				\addplot[only marks,mark=*, mark size=1.5pt,black] coordinates {(1/2, 1/2)};
				\nextgroupplot[title={\footnotesize Select POH}]
				\addplot[thick,patch,mesh,draw,black,patch type=rectangle,line width=0.3mm] coordinates {(0,0) (1,0) (1,1) (0,1)};
				\draw [black, thick, mark size=0.05pt, fill=green!40,opacity=0.4,line width=0.3mm] (axis cs:0,0) rectangle (axis cs:1,1);
				\addplot[only marks,mark=*, mark size=1.5pt,black] coordinates {(1/2, 1/2)};
				\nextgroupplot[title={\footnotesize Sample}]
				\addplot[thick,patch,mesh,draw,black,patch type=rectangle,line width=0.3mm] coordinates {(0,0) (1,0) (1,1) (0,1)};
				\draw [black, thick, mark size=0.05pt, fill=green!40,opacity=0.4,line width=0.3mm] (axis cs:0,0) rectangle (axis cs:1,1);
				\addplot[only marks,mark=*, mark size=1.5pt,black] coordinates {(1/2, 1/2)};
				\addplot[only marks,mark=*, mark size=1.5pt,blue] coordinates {(1/6, 1/2) (5/6, 1/2) (1/2, 1/6) (1/2, 5/6)};
				\nextgroupplot[title={\footnotesize Subdivide}, ylabel={Iteration $1$}, ylabel style={at={(axis description cs:1.2,0.975)},anchor=west,rotate=180,draw, rectangle, rounded corners, inner sep=3pt},]
				\addplot[thick,patch,mesh,draw,black,patch type=rectangle,line width=0.3mm] coordinates {(0,0) (1,0) (1,1) (0,1)};
				\draw [black, thick, mark size=0.05pt, fill=green!40,opacity=0.4,line width=0.3mm] (axis cs:0,0) rectangle (axis cs:1,1);
				\addplot[thick,patch,mesh,draw,black,patch type=rectangle,line width=0.3mm] coordinates {(1/3, 0) (1/3, 1) (2/3, 1) (2/3, 0)};
				\addplot[thick,patch,mesh,draw,black,patch type=rectangle,line width=0.3mm] coordinates {(1/3, 1/3) (1/3, 2/3) (2/3, 2/3) (2/3, 1/3)};
				\addplot[only marks,mark=*, mark size=1.5pt,black] coordinates {(1/2, 1/2)};
				\addplot[only marks,mark=*, mark size=1.5pt,blue] coordinates {(1/6, 1/2) (5/6, 1/2) (1/2, 1/6) (1/2, 5/6)};
				\nextgroupplot
				\addplot[thick,patch,mesh,draw,black,patch type=rectangle,line width=0.3mm] coordinates {(0,0) (1,0) (1,1) (0,1)};
				\addplot[thick,patch,mesh,draw,black,patch type=rectangle,line width=0.3mm] coordinates {(1/3, 0) (1/3, 1) (2/3, 1) (2/3, 0)};
				\addplot[thick,patch,mesh,draw,black,patch type=rectangle,line width=0.3mm] coordinates {(1/3, 1/3) (1/3, 2/3) (2/3, 2/3) (2/3, 1/3)};
				\addplot[only marks,mark=*, mark size=1.5pt,black] coordinates {(1/2, 1/2) (1/6, 1/2) (5/6, 1/2) (1/2, 1/6) (1/2, 5/6)};
				\nextgroupplot
				\addplot[thick,patch,mesh,draw,black,patch type=rectangle,line width=0.3mm] coordinates {(0,0) (1,0) (1,1) (0,1)};
				\draw [black, thick, mark size=0.05pt, fill=green!40,opacity=0.4,line width=0.3mm] (axis cs:0,0) rectangle (axis cs:1/3,1);
				\addplot[thick,patch,mesh,draw,black,patch type=rectangle,line width=0.3mm] coordinates {(1/3, 0) (1/3, 1) (2/3, 1) (2/3, 0)};
				\addplot[thick,patch,mesh,draw,black,patch type=rectangle,line width=0.3mm] coordinates {(1/3, 1/3) (1/3, 2/3) (2/3, 2/3) (2/3, 1/3)};
				\addplot[only marks,mark=*, mark size=1.5pt,black] coordinates {(1/2, 1/2) (1/6, 1/2) (5/6, 1/2) (1/2, 1/6) (1/2, 5/6)};
				\nextgroupplot
				\addplot[thick,patch,mesh,draw,black,patch type=rectangle,line width=0.3mm] coordinates {(0,0) (1,0) (1,1) (0,1)};
				\draw [black, thick, mark size=0.05pt, fill=green!40,opacity=0.4,line width=0.3mm] (axis cs:0,0) rectangle (axis cs:1/3,1);
				\addplot[thick,patch,mesh,draw,black,patch type=rectangle,line width=0.3mm] coordinates {(1/3, 0) (1/3, 1) (2/3, 1) (2/3, 0)};
				\addplot[thick,patch,mesh,draw,black,patch type=rectangle,line width=0.3mm] coordinates {(1/3, 1/3) (1/3, 2/3) (2/3, 2/3) (2/3, 1/3)};
				\addplot[only marks,mark=*, mark size=1.5pt,black] coordinates {(1/2, 1/2) (1/6, 1/2) (5/6, 1/2) (1/2, 1/6) (1/2, 5/6)};
				\addplot[only marks,mark=*, mark size=1.5pt,blue] coordinates {(1/6, 1/6) (1/6, 5/6)};
				\nextgroupplot[ylabel={Iteration $2$}, ylabel style={at={(axis description cs:1.2,0.975)},anchor=west,rotate=180,draw, rectangle, rounded corners, inner sep=3pt},]
				\addplot[thick,patch,mesh,draw,black,patch type=rectangle,line width=0.3mm] coordinates {(0,0) (1,0) (1,1) (0,1)};
				\draw [black, thick, mark size=0.05pt, fill=green!40,opacity=0.4,line width=0.3mm] (axis cs:0,0) rectangle (axis cs:1/3,1);
				\addplot[thick,patch,mesh,draw,black,patch type=rectangle,line width=0.3mm] coordinates {(1/3, 0) (1/3, 1) (2/3, 1) (2/3, 0)};
				\addplot[thick,patch,mesh,draw,black,patch type=rectangle,line width=0.3mm] coordinates {(0, 1/3) (0, 2/3) (2/3, 2/3) (2/3, 1/3)};
				\addplot[only marks,mark=*, mark size=1.5pt,black] coordinates {(1/2, 1/2) (1/6, 1/2) (5/6, 1/2) (1/2, 1/6) (1/2, 5/6)};
				\addplot[only marks,mark=*, mark size=1.5pt,blue] coordinates {(1/6, 1/6) (1/6, 5/6)};
				\nextgroupplot
				\addplot[thick,patch,mesh,draw,black,patch type=rectangle,line width=0.3mm] coordinates {(0,0) (1,0) (1,1) (0,1)};
				\addplot[thick,patch,mesh,draw,black,patch type=rectangle,line width=0.3mm] coordinates {(1/3, 0) (1/3, 1) (2/3, 1) (2/3, 0)};
				\addplot[thick,patch,mesh,draw,black,patch type=rectangle,line width=0.3mm] coordinates {(0, 1/3) (0, 2/3) (2/3, 2/3) (2/3, 1/3)};
				\addplot[only marks,mark=*, mark size=1.5pt,black] coordinates {(1/2, 1/2) (1/6, 1/2) (5/6, 1/2) (1/2, 1/6) (1/2, 5/6) (1/6, 1/6) (1/6, 5/6)};
				\nextgroupplot
				\addplot[thick,patch,mesh,draw,black,patch type=rectangle,line width=0.3mm] coordinates {(0,0) (1,0) (1,1) (0,1)};
				\draw [black, thick, mark size=0.05pt, fill=green!40,opacity=0.4,line width=0.3mm] (axis cs:2/3,0) rectangle (axis cs:1,1);
				\draw [black, thick, mark size=0.05pt, fill=green!40,opacity=0.4,line width=0.3mm] (axis cs:0,0) rectangle (axis cs:1/3,1/3);
				\addplot[thick,patch,mesh,draw,black,patch type=rectangle,line width=0.3mm] coordinates {(1/3, 0) (1/3, 1) (2/3, 1) (2/3, 0)};
				\addplot[thick,patch,mesh,draw,black,patch type=rectangle,line width=0.3mm] coordinates {(0, 1/3) (0, 2/3) (2/3, 2/3) (2/3, 1/3)};
				\addplot[only marks,mark=*, mark size=1.5pt,black] coordinates {(1/2, 1/2) (1/6, 1/2) (5/6, 1/2) (1/2, 1/6) (1/2, 5/6) (1/6, 1/6) (1/6, 5/6)};
				\nextgroupplot
				\addplot[thick,patch,mesh,draw,black,patch type=rectangle,line width=0.3mm] coordinates {(0,0) (1,0) (1,1) (0,1)};
				\draw [black, thick, mark size=0.05pt, fill=green!40,opacity=0.4,line width=0.3mm] (axis cs:2/3,0) rectangle (axis cs:1,1);
				\draw [black, thick, mark size=0.05pt, fill=green!40,opacity=0.4,line width=0.3mm] (axis cs:0,0) rectangle (axis cs:1/3,1/3);
				\addplot[thick,patch,mesh,draw,black,patch type=rectangle,line width=0.3mm] coordinates {(1/3, 0) (1/3, 1) (2/3, 1) (2/3, 0)};
				\addplot[thick,patch,mesh,draw,black,patch type=rectangle,line width=0.3mm] coordinates {(0, 1/3) (0, 2/3) (2/3, 2/3) (2/3, 1/3)};
				\addplot[only marks,mark=*, mark size=1.5pt,black] coordinates {(1/2, 1/2) (1/6, 1/2) (5/6, 1/2) (1/2, 1/6) (1/2, 5/6) (1/6, 1/6) (1/6, 5/6)};
				\addplot[only marks,mark=*, mark size=1.5pt,blue] coordinates {(5/6, 1/6) (5/6, 5/6) (1/18, 1/6) (5/18, 1/6) (1/6, 5/18) (1/6, 1/18)};
				\nextgroupplot[ylabel={Iteration $3$}, ylabel style={at={(axis description cs:1.2,0.975)},anchor=west,rotate=180,draw, rectangle, rounded corners, inner sep=3pt},]
				\addlegendimage{only marks,mark=*,color=black, mark size=3pt}
				\addlegendentry{Sampled point}
				\addlegendimage{only marks,mark=*,color=blue, mark size=3pt}
				\addlegendentry{New sampling point}
				\addlegendimage{area legend,fill=green!40,opacity=0.4,line width=0.3mm}
				\addlegendentry{Selected POH}
				\addlegendimage{area legend,black,fill=white,opacity=0.5}
				\addlegendentry{Unselected region}
				\addplot[thick,patch,mesh,draw,black,patch type=rectangle,line width=0.3mm] coordinates {(0,0) (1,0) (1,1) (0,1)};
				\draw [black, thick, mark size=0.05pt, fill=green!40,opacity=0.4,line width=0.3mm] (axis cs:2/3,0) rectangle (axis cs:1,1);
				\draw [black, thick, mark size=0.05pt, fill=green!40,opacity=0.4,line width=0.3mm] (axis cs:0,0) rectangle (axis cs:1/3,1/3);
				\addplot[thick,patch,mesh,draw,black,patch type=rectangle,line width=0.3mm] coordinates {(1/3, 0) (1/3, 1) (2/3, 1) (2/3, 0)};
				\addplot[thick,patch,mesh,draw,black,patch type=rectangle,line width=0.3mm] coordinates {(0, 1/3) (0, 2/3) (1, 2/3) (1, 1/3)};
				\addplot[thick,patch,mesh,draw,black,patch type=rectangle,line width=0.3mm] coordinates {(0, 1/9) (0, 2/9) (1/3, 2/9) (1/3, 1/9)};
				\addplot[thick,patch,mesh,draw,black,patch type=rectangle,line width=0.3mm] coordinates {(1/9, 1/9) (1/9, 2/9) (2/9, 2/9) (2/9, 1/9)};
				\addplot[only marks,mark=*, mark size=1.5pt,black] coordinates {(1/2, 1/2) (1/6, 1/2) (5/6, 1/2) (1/2, 1/6) (1/2, 5/6) (1/6, 1/6) (1/6, 5/6)};
				\addplot[only marks,mark=*, mark size=1.5pt,blue] coordinates {(5/6, 1/6) (5/6, 5/6) (1/18, 1/6) (5/18, 1/6) (1/6, 5/18) (1/6, 1/18)};
			\end{groupplot}
	\end{tikzpicture}}
	\caption{Visualization of selection, sampling, and subdivision in \direct{} algorithm on the two-dimensional problem.}
	\label{fig_examplemain}
\end{figure}

The \direct{} algorithm~\cite{Jones1993_DIRECT} is a deterministic iterative global optimization algorithm that involves four major steps: \textit{initialization}, \textit{selection}, \textit{sampling}, and \textit{Subdivision}. 
Once the \textit{initialization} step is completed, the algorithm selects Potential Optimal Hyper-rectangles (POHs), samples, evaluates, and trisects them in subsequent iterations. 
Most \direct-type extensions and modifications follow the same algorithmic framework, which is summarized in Algorithm~\ref{alg_algorithm}, while \Cref{fig_examplemain} illustrates the process of the algorithm in its initial three iterations on a two-variable problem.
We will briefly review each algorithm step in the following paragraphs.

\begin{algorithm}[ht!]
	\footnotesize
	\nonl\textbf{Input:} \\
	\nonl \quad $f$: Objective function\;
	\nonl \quad $\mathcal{X}$: Decision space\;
	\nonl \quad \textit{OPT}: structure with optimization options\;
	\nonl\textbf{Output:} \\
	\nonl \quad  $f^{\rm min}_k$, $\mathbf{c}^{\rm min}_k$: Optimal solution and its corresponding point\; 
	\nonl \quad $t$ (time), $k$ (iterations), $m$ (function evaluations): Performance measures\;
	\algrule
	\textbf{Initialization}: Normalize $\mathcal{X}$ to $\bar{\mathcal{X}}$. Evaluate $f$ at the center point $\mathbf{c}^1$. Set $f^{\rm min}_1 \leftarrow f(\mathbf{c}^1)$ and $c^{\rm min}_1 \leftarrow \mathbf{c}^1$. Initialize $t$, $k \leftarrow 1$, $m \leftarrow 1$, and \textit{stopping criteria}\; \label{alg:initialization}
	\While{stopping criteria \textnormal{are not satisfied}}{
		\textbf{Selection}: Identify the set $\mathcal{S}_k \subseteq \mathcal{P}_k$ of POHs\; \label{alg:selection}
		\textbf{Sampling}: For each $\bar{\mathcal{X}}^j_k \in \mathcal{S}_k$ evaluate $f$ at newly sampled points\; \label{alg:sampling}
		\textbf{Subdivision}: Each $\bar{\mathcal{X}}^j_k \in \mathcal{S}_k$ subdivide (trisect) along all long sides\; \label{alg:subdivision}
		Set $k \leftarrow k + 1$ and update $\mathcal{P}_k$, $f^{\rm min}_k$, $\mathbf{c}^{\rm min}_k$, $m$, $t$\;
	}
	\textbf{Return}: $f^{\rm min}_k, \mathbf{c}^{\rm min}_k$, and performance measures ($t$, $k$, $m$).
	\caption{Main steps of \direct-type algorithms}
	\label{alg_algorithm}
\end{algorithm}

\paragraph{Initialization}
In the \textit{Initialization} step, the \direct{} algorithm normalizes the feasible region $\mathcal{X}$ to the unit hyper-cube $\bar{\mathcal{X}}$. 
It only refers to the original space $\mathcal{X}$ when evaluating the objective function $f$. 
During this step, the algorithm assesses the objective function at the midpoint $\mathbf{c}^1 \in \bar{\mathcal{X}}^1_1$ and initiates the formation of the partition of $\mathcal{P}$. 
At iteration $k$, this partition is defined as:
\begin{equation}
	\label{eq_partition_set}
	\mathcal{P}_k = \{ \bar{\mathcal{X}}^i_k : i \in \mathbb{I}_k \},
\end{equation}
where
\begin{equation}
	\label{eq:rectangle}
	\bar{\mathcal{X}}^i_k = [\mathbf{l}_k^i, \mathbf{u}_k^i] = \{ \mathbf{x} \in \bar{\mathcal{X}}: 0 \leq l_{k_j}^i \leq x_j \leq u_{k_j}^i \leq 1, j = 1,\dots, n, \forall i \in \mathbb{I}_k \},
\end{equation}
and $\mathbb{I}_k$ is the index set identifying the current partition $ \mathcal{P}_k $. 
The subsequent partition, $\mathcal{P}_{k+1}$, is obtained through the subdivision of the selected POHs from the current partition $\mathcal{P}_k$.

\paragraph{Selection}
The \textit{Selection} procedure at the first iteration is straightforward, as there is only one candidate $\bar{\mathcal{X}}^1_1 \in \mathcal{P}_1$. 
To identify the POHs in subsequent iterations, the algorithm uses lower-bound estimates of the objective function $f$ over each hyper-rectangle in the current partition, as formalized in \Cref{def_potOptRect}.

\begin{definition}[Potentially optimal hyper-rectangle~\cite{Jones1993_DIRECT}]
	\label{def_potOptRect}
	Let $\mathbf{c}^i$ denote the center sample point of hyper-rectangle $\bar{\mathcal{X}}_k^i$, and let $\delta_k^i$ denote its measure. 
	Assume a positive constant $\varepsilon_{\rm DIR} > 0$, and let $f_k^{\rm min}$ denote the best objective value obtained up to iteration $k$.
	
	A hyper-rectangle $\bar{\mathcal{X}}_k^j$, with $j \in \mathbb{I}_k$, is said to be \emph{potentially optimal} if there exists a Lipschitz constant $\tilde{L} > 0$ such that
	\begin{eqnarray}
		f(\mathbf{c}^j) - \tilde{L}\delta_k^j & \leq & f(\mathbf{c}^i) - \tilde{L}\delta_k^i, \quad \forall i \in \mathbb{I}_k, \label{eqn_potOptRect1} \\
		f(\mathbf{c}^j) - \tilde{L}\delta_k^j & \leq & f_k^{\rm min} - \varepsilon_{\rm DIR}, \label{eqn_potOptRect2}
	\end{eqnarray}
	where the size of hyper-rectangle $\bar{\mathcal{X}}_k^i$ is defined by
	\begin{equation}
		\label{eq:distance}
		\delta_k^i = \frac{1}{2}\|\mathbf{u}_k^i - \mathbf{l}_k^i\|_2.
	\end{equation}
\end{definition}

A hyper-rectangle $\bar{\mathcal{X}}_k^j$ is therefore considered potentially optimal if, for some positive constant $\tilde{L}$, its lower Lipschitz bound given by the left-hand side of \eqref{eqn_potOptRect1} is not greater than that of any other hyper-rectangle in the current partition $\mathcal{P}_k$. 
Condition \eqref{eqn_potOptRect2} introduces the parameter $\varepsilon_{\rm DIR}$ to avoid excessive refinement near already identified local minima.

\begin{remark}
	The selection of the POH is the most studied and refined step in the literature, indicating its importance. 
	Many different strategies have been proposed to improve the original selection, including methods to control the $\varepsilon_{\rm DIR}$ parameter~\cite{Finkel2004_DIRECT_restart}, the selection at various levels~\cite{Liu2015_MrDIRECT} and with multiple $\varepsilon_{\rm DIR}$ parameters~\cite{Liu2017_MrDIRECT}, using different norms in \eqref{eq:distance}~\cite{Gablonsky2001_DIRECTl}, and replacing $f^{\rm min}_k$ with the median~\cite{Finkel2006_DIRECT_m} or average~\cite{Liu2013_DIRECT_a} function values.
	Also, there can be a significant number of tied values within the same hyper-rectangle measure, which issue has been tackled in~\cite{Gablonsky2001_DIRECTl, Jones2001_DIRECT_REV}.
	Some authors have even integrated completely different selection schemes, such as aggressive~\cite{Baker2000_Aggressive_direct}, Pareto~\cite{Mockus2011_Plo}, and reduced Pareto~\cite{Mockus2017_Plor}. 
	Recent studies~\cite{Stripinis2024_I_DTC_GL, Stripinis2024_HALRECT} suggest that the two-step global and local Pareto selection scheme~\cite{Stripinis2018_DIRECT_GL} is the most suitable and efficient selection that is currently available. 
\end{remark}

\paragraph{Partitioning procedure}
The following steps in the \direct{} iteration involve \textit{sampling} and \textit{subdivision}. 
These can be thought of as the partitioning procedure of the partition $\mathcal{P}_k$. 
This procedure determines which patterns the $\bar{\mathcal{X}}$ will be subdivided into and where the samples will be taken within these patterns.

\subparagraph{Sampling}
When a POH ($\bar{\mathcal{X}}^i_k$) is selected, \direct{} samples points at:
\begin{equation}
	\label{eq_newpoints}
	\mathbf{c}^i \pm \bar{d}^i_k\mathbf{e}_j, \quad j = \mathbb{J}^i_k,
\end{equation}
where $j$ belongs to the set $\mathbb{J}^i_k$. 
These points are sampled along the sides of the POH, with $\bar{d}^i_k$ equaling one-third of the maximum side length of $\bar{\mathcal{X}}^i_k$, and $\mathbf{e}_j$ is the $j$th Euclidean base-vector. 
Initially, the proposed \direct{} samples all sides with the maximum side lengths. 
However, only two additional points must be sampled along the longest side. 
The number of samples per POH can vary from $2$ to $2n$, depending on the number of sides that share the maximum length.

\subparagraph{Subdivision}
The division process in \direct{} involves dividing the hyper-rectangle into three equal parts, along the longest sides, in an $n$-dimensional space. 
If multiple sides share the same length, the trisection process begins with the sides with the smallest value of $w^i_j$ and moves to the highest value. 
The $w^i_j$ is determined as the minimum value of the function sampled along the dimension $j$:
\begin{equation}
	\label{eq_w}
	w^i_j = \min \left\{ f(\mathbf{c}^i + \bar{d}^i_k\mathbf{e}_j), f(\mathbf{c}^i - \bar{d}^i_k\mathbf{e}_j) \right\}, \quad j \in \mathbb{J}^i_k.
\end{equation}
The centers of the newly created ``left'' and ``right'' hyper-rectangles take newly sampled points, while the original center point becomes the center of the middle hyper-rectangle. 
The \direct{} algorithm's partitioning strategy aims to ensure that the best function values are contained in the largest hyper-rectangles. 
This strategy helps the algorithm explore points close to favorable function values, emphasizing local search while maintaining global search capabilities.

\begin{remark}
	Partitioning procedures that can be used with the \direct{} algorithm have been the subject of extensive research.
	Six other partitioning schemes have been proposed for use with \direct-type algorithms~\cite{Jones2001_DIRECT_REV, Stripinis2024_HALRECT, Paulavicius2018_BIRECT, Sergeyev2006_ADC, Guessoum2023_BIRECT_V, Paulavicius2013_DISIMPL}. 
	While, the original strategy suggested that the hyper-rectangles should be partitioned according to all the longest side lengths. 
	All the newly proposed schemes have restricted this partitioning to only the longest side length. 
	Although various studies~\cite{Stripinis2024_I_DTC_GL, Stripinis2024_HALRECT} show the advantages of one strategy over the other, selecting the appropriate algorithmic composition highly depends on the use case.
\end{remark}

\paragraph{Example of the optimization process with \textnormal{\direct}}

\Cref{ex11} clarifies the performance issues of the \textnormal{\direct} and some altered versions that help to mitigate them using a simple linear test problem.

\begin{example}
	\label{ex11}
	The left side of \Cref{fig_poh}, illustrates the performance of the \textnormal{\direct} algorithm on a simple two-dimensional ($n=2$) linear function:
	\begin{equation}
		\label{eq_prob1}
		\begin{aligned}
			& \min_{\mathbf{x}\in [\mathbf{0}, \mathbf{1}]} && f(\mathbf{x}) = 1 + \sum_{i=1}^{n} x_i
		\end{aligned}
	\end{equation}
	The global minimum value of \eqref{eq_prob1} is $f^* = 1$ at the point $\mathbf{x}^* = (0, 0)$. 
	For the experiment, we set the function target value within the absolute function value error of $10^{-4}$. 
	The original \direct{} algorithm required $613$ function evaluations to solve this simple problem.
	
	A recent review paper~\cite{Jones2021_review} by the original algorithm's author demonstrated how \textnormal{\direct}-type algorithms are affected by convergence and dimensionality issues when finding the solution within varying $n$, using the same problem~\eqref{eq_prob1}. 
	To improve the convergence speed, the authors investigated existing variations and suggested the combination (let's call it \textnormal{\directc}) from three studies~\cite{Jones1993_DIRECT, Gablonsky2001_DIRECTl, Jones2001_DIRECT_REV}, which reduced the number of function evaluations needed to solve the problem from $613$ to $105$ (see middle part of \Cref{fig_poh}). 
	The performance could be further improved by substituting the selection of the POH scheme with \cite{Mockus2017_Plor}, which gives equal attention to the exploitation and exploration phases. 
	We incorporated this selection in the suggested framework, let's call it \textnormal{\plorc}, and we further can reduce the number of function evaluations needed to solve the problem to 69, as illustrated on the left side of \Cref{fig_poh}.
	But this is perhaps the cheapest version of the \textnormal{\direct}-type algorithm that could be designed today.
	\begin{figure}[ht!]
		\centering
		\includegraphics[width=1\linewidth]{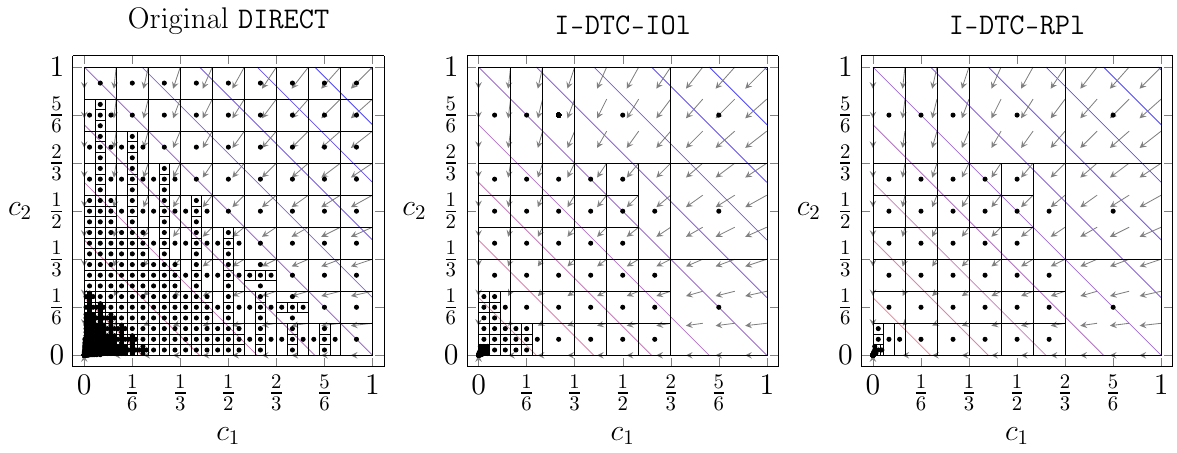}
		\caption{Visualization of three \direct-type algorithms design spaces $D$ solving two-dimensional linear test problem \eqref{eq_prob1}.}
		\label{fig_poh}
	\end{figure}
\end{example}

\begin{problem}
	Reducing the exploration capabilities, indeed, can mitigate the curse of dimensionality. However, insufficient global search is a clear risk of spending excessive function evaluations tuning suboptimal solutions, thus delaying finding the global minimum~\cite{Jones2021_review, Stripinis2018_DIRECT_GL}.
\end{problem}

\subsection{Hybrid \direct-type algorithms}

While different practical applications may require specific hybridization strategies~\cite{Stripinis2024_review}, in this work, we summarize six popular and useful hybrid \direct-type algorithms in \Cref{tab_halgorithms}.

\begin{table}[ht!]
	\caption{Summary of existing hybrid \direct-type algorithmic frameworks and their combinations}
	\label{tab_halgorithms}
	\resizebox{\textwidth}{!}{%
		\begin{tabular}{l|C{2.5cm}|C{2.5cm}|C{2.5cm}|C{2.5cm}|C{2.5cm}|C{2.5cm}}
			\toprule
			\textbf{Step/Algorithm} & \birmin & \directrev & \glccluster & \dirmin & \directt & \directsqp \\
			\midrule
			Year and reference & $2020$~\cite{Paulavicius2020_BIRMIN} & $2001$~\cite{Jones2001_DIRECT_REV} & $2004$~\cite{Holmstrom2004} & $2010$~\cite{Liuzzi2010_DIRMIN} & $2023$~\cite{Kanayama2023_tDIRECT} & $2014$~\cite{Li2016_DIRECTsqp} \\
			\midrule
			\textit{Selection} scheme & \multicolumn{6}{c}{Utilizes \Cref{def_potOptRect} to find the set of POHs} \\
			\midrule
			\multirow{2}{3cm}{\textit{Selection} improvement} & \multicolumn{3}{c|}{Breaks $\bar{\mathcal{X}}_k^i$ ties at each $\delta$} & \multicolumn{3}{c}{\multirow{2}{*}{--}} \\
			\cmidrule{2-4}
			& globally-biased & \multicolumn{2}{c|}{--} & \multicolumn{3}{c}{}  \\
			\midrule
			\multirow{2}{*}{\textit{Sampling}} & Two points on & \multicolumn{5}{c}{\multirow{2}{*}{Midpoints of each $\bar{\mathcal{X}}_k^i$}} \\
			& $\bar{\mathcal{X}}_k^i$ diagonal \\
			\midrule
			\multirow{2}{*}{\textit{Subdivision}} & Bisects one & \multicolumn{2}{c|}{\multirow{2}{*}{Trisects one longest side}} & \multicolumn{3}{c}{\multirow{2}{*}{Trisects all longest sides}} \\
			& longest side & \multicolumn{2}{c|}{} \\
			\midrule
			\multirow{2}{3cm}{\textit{Hill-climbing} algorithm} & \texttt{interior} & \multirow{2}{*}{\texttt{Unspecified}} & \multirow{2}{*}{\texttt{NPSOL}} & \multirow{2}{*}{\texttt{NMonNC}} & \texttt{conjugate} & \multirow{2}{*}{\texttt{SQP}} \\
			& \texttt{point} &&&& \texttt{gradient} &   \\
			\midrule
			\multirow{3}{3cm}{Run hill-climbers starting at:} & \multicolumn{2}{c|}{\multirow{3}{*}{$\mathbf{c}^{\rm min}_k$, if \direct{} improves $f^{\rm min}_k$}} & \multirow{3}{2.25cm}{\centering best $\mathbf{x}$ from each cluster} & \multicolumn{3}{c}{$\mathbf{c}^i$ from every:} \\
			\cmidrule{5-7}
			& \multicolumn{2}{c|}{} &  & \multirow{2}{*}{$\bar{\mathcal{X}}_k^i \in S_k$} & \multirow{2}{*}{$\bar{\mathcal{X}}_k^i \in P_k$} & \multirow{2}{2.25cm}{\centering $\bar{\mathcal{X}}_k^i \in S_k$, if $\delta_k^i \leq \delta^{\rm limit}$} \\
			& \multicolumn{2}{c|}{} &&&& \\
			\bottomrule
	\end{tabular}}
\end{table}

Three of the six hybrid algorithms (\dirmin, \directt, and \directsqp) use the original \direct{} algorithmic framework in combination with \textit{hill-climbing} optimizers. 
Two of these algorithms, \dirmin{} and \directt, excessively use hill-climbers, starting from each POH (\dirmin) or from every sampled point (\directt). 
The \directsqp{} algorithm uses hill-climber more cautiously.
Motivated by the fact that the global optima can be located far from the starting points, the authors suggested executing the hill-climber beginning from the POHs only if it reaches some prescribed limit ($\delta^{\rm limit}$).

Other researchers have combined hill-climbers and \direct{} more carefully.
Other authors modified the selection and partitioning strategies for their approaches.
In particular, \birmin{} and \directrev{} mostly rely on \direct{} search and use hill-climbers only if there is some improvement in their search. 
The \glccluster{} algorithm uses an entirely different approach to utilizing hill-climber. 
The algorithm starts the hill-climber from the best point of each adaptively clustering algorithm-made cluster.

\section{Description of the Proposed Approach}
\label{sec_algorithm}

In this section, we present a description of the newly proposed \xdtcgl{} algorithm, which is an extension of the \idtcgl.

\subsection{From \idtcgl{} to \xdtcgl}

\subsubsection{\idtcgl{} Algorithm}

The \idtcgl{} algorithm~\cite{Stripinis2024_I_DTC_GL} employs a partitioning strategy that differs slightly from the original \direct{} algorithm. 
It uses a hyper-rectangular partitioning approach based on $1-$\texttt{D}imensional \texttt{T}risection, and objective function evaluations are conducted at \texttt{C}enter points (\texttt{I-DTC})~\cite{Jones2001_DIRECT_REV}. 
Instead of subdividing all the longest sides of the hyper-rectangle, the algorithm selects only one with the least splits over the entire algorithm search.
Additionally, the algorithm utilizes a two-step-based (\texttt{G}lobal-\texttt{L}ocal) Pareto selection (\texttt{GL}) scheme~\cite{Stripinis2021_pDIRECT_GLce, Stripinis2018_DIRECT_GL}, which is the most effective for complex problems based on multiple studies~\cite{Stripinis2024_HALRECT, Stripinis2022_DIRECTGO}.

\subsubsection{Implementing the Dynamic Partitioning Strategy Into \xdtcgl}

When the algorithm begins to iteratively tune the solution, we need to be cautious of the potentially costly global drag that can occur. 
Instead of relying on a fixed subdivision scheme, the proposed approach adaptively refines a selected POH by constructing a one-dimensional surrogate model along its longest (i.e., least partitioned) side.
Based on this model, the predicted minimum is estimated, and the sub-hyper-rectangle containing this predicted minimum is further subdivided if an improvement in $f^{\min}$ is expected. 
Unlike the original one-shot partitioning, this process can be repeated, allowing the initially selected hyper-rectangle to be subdivided multiple times through its most promising sub-hyper-rectangles, each time along the longest side, until no further improvement is indicated.
This allows the solution region to be progressively tightened, as illustrated later in~\Cref{fig_example1}.

\begin{figure}[ht!]
	\centering
	\resizebox{0.95\textwidth}{!}{%
		\begin{tikzpicture}
			\begin{groupplot}[group style={group size=2 by 1, horizontal sep=0pt}, ymin=-1.8, ymax=5.5, xmin=-1.7, xmax=7, width=0.5\textwidth, height=0.375\textwidth, axis line style={draw=none}, tick style={draw=none},yticklabel=\empty,xticklabel=\empty]
				\nextgroupplot[title={\small Quadratic}]
				\draw[->] (-1.5,0) -- (6.5,0) node[below] {\scriptsize $x$};
				\draw[->] (-0.5,-1) -- (-0.5,4.5) node[left] {\scriptsize $f(x)$};
				\draw[domain=0:6, smooth, variable=\x, thick, black!50, dashed] plot (\x, {0.375*\x*\x - 2*\x + 3.625}) node[left] {\scriptsize $\hat{f}(c)$}; 
				\filldraw[black] (1,2) circle (1.5pt) node[left] {};
				\filldraw[black] (3,1) circle (1.5pt) node[above] {};
				\filldraw[black] (5,3) circle (1.5pt) node[right] {};
				\filldraw[black] (1,0) circle (1.5pt) node[below] {\scriptsize $c^{\rm left}$};
				\filldraw[black] (3,0) circle (1.5pt) node[below] {\scriptsize $c^{\rm middle}$};
				\filldraw[black] (5,0) circle (1.5pt) node[below] {\scriptsize $c^{\rm right}$};
				\filldraw[black!50] (8/3,0) circle (1.5pt) node[below] {};
				\filldraw[black!50] (8/3,0.95) circle (1.5pt) node[above] {\scriptsize $\hat{f}^{\rm min}$};
				\draw[black] (2,-0.1) -- (2,0.1);
				\draw[black] (6,-0.1) -- (6,0.1);
				\draw[black] (4,-0.1) -- (4,0.1);
				\draw[black] (0,-0.1) -- (0,0.1);
				\draw[dotted, thick, smooth, black!50] (8/3,0.95) -- (8/3,0);
				\draw[dotted, thick, smooth, black] (1,0) -- (1,2);
				\draw[dotted, thick, smooth, black] (3,0) -- (3,1);
				\draw[dotted, thick, smooth, black] (5,0) -- (5,3);
				\draw[dotted, thick, smooth, black] (0,-1) -- (6,-1);
				\draw[black] (6,-0.8) -- (6,-1.2);
				\draw[black] (0,-0.8) -- (0,-1.2);
				\node at (3,-1.45) {\scriptsize $\bar{\mathcal{X}}^1_1$};
				\nextgroupplot[title={\small Linear}]
				\draw[->] (-1.5,0) -- (6.5,0) node[below] {\scriptsize $x$};
				\draw[->] (-0.5,-1) -- (-0.5,4.5) node[left] {\scriptsize $f(x)$};
				\draw[smooth, thick, black!50, dashed] (2,0) -- (6,4) node[above] {\scriptsize $\hat{f}(c)$}; 
				\draw[smooth, thick, black!50, dashed] (4,0.5) -- (0,2.5) node[above] {\scriptsize $\hat{f}(c)$};
				\filldraw[black!50] (2,0) circle (1pt) node[above] {\scriptsize $\hat{f}^{\rm min}$};
				\filldraw[black] (1,2) circle (1.5pt) node[left] {};
				\filldraw[black] (3,1) circle (1.5pt) node[above] {};
				\filldraw[black] (5,3) circle (1.5pt) node[right] {};
				\filldraw[black] (1,0) circle (1.5pt) node[below] {\scriptsize $c^{\rm left}$};
				\filldraw[black] (3,0) circle (1.5pt) node[below] {\scriptsize $c^{\rm middle}$};
				\filldraw[black] (5,0) circle (1.5pt) node[below] {\scriptsize $c^{\rm right}$};
				\draw[black] (2,-0.1) -- (2,0.1);
				\draw[black] (6,-0.1) -- (6,0.1);
				\draw[black] (4,-0.1) -- (4,0.1);
				\draw[black] (0,-0.1) -- (0,0.1);
				\draw[dotted, thick, smooth, black] (1,0) -- (1,2);
				\draw[dotted, thick, smooth, black] (3,0) -- (3,1);
				\draw[dotted, thick, smooth, black] (5,0) -- (5,3);
				\draw[dotted, thick, smooth, black] (0,-1) -- (6,-1);
				\draw[black] (6,-0.8) -- (6,-1.2);
				\draw[black] (0,-0.8) -- (0,-1.2);
				\node at (3,-1.45) {\scriptsize $\bar{\mathcal{X}}^1_1$};
			\end{groupplot}
	\end{tikzpicture}}
	\caption{Quadratic (left side) and linear (right side) approximation using three-point interpolation in subdivision of the first hyper-rectangle ($\bar{\mathcal{X}}^1_1$) in \direct.}
	\label{fig_aprox}
\end{figure}
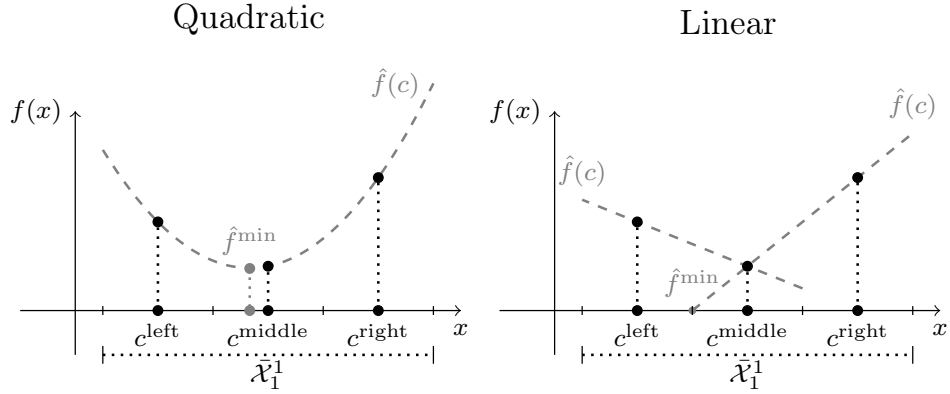

To explore the potential in each POH, we investigated two different approximations, quadratic and linear, as shown in \Cref{fig_aprox}.
The left panel of the figure illustrates the quadratic approximation, incorporating all three points sampled subsequent to the subdivision of the $\bar{\mathcal{X}}^1_1$ hyper-rectangle along its longest side. 
Conversely, the right panel depicts the linear approximation utilizing two pairs of points. 
The search for the minimum is conducted within the bounds of the considered $\bar{\mathcal{X}}^1_1$ hyper-rectangle. 
In both cases, the middle hyper-rectangle encloses the $\hat{f}^{\rm min}$ value, with the linear approximation predicting better enhancement in the solution.

The formal definition of dynamic partitioning is given in \Cref{def_dynPartRule}.

\begin{definition}[Dynamic partitioning]
	\label{def_dynPartRule}
	Let $\bar{\mathcal{X}}_k^i$ denote the selected POH at iteration $k$, where $i$ is its index in the current partition $\mathcal{P}_k$. Its bounds are given by $[\mathbf{l}_k^i, \mathbf{u}_k^i]$, and its center is denoted by $\mathbf{c}^i$.
	
	Let $m$ denote the total number of function evaluations performed up to iteration $k$. Two new sample points, indexed by $m+1$ and $m+2$, are generated along coordinate $j$ at distance $\bar{d}_k^i$ from the center coordinate $c_j^i$, where $\bar{d}_k^i$ is equal to one-third of the maximal side length of $\bar{\mathcal{X}}_k^i$. Denote the corresponding coordinate values by $c_j^{m+1}$ and $c_j^{m+2}$.
	
	Let $\hat{f}^{\rm min}$ and $\hat{c}^{\rm min}$ denote, respectively, the approximate minimum value and its corresponding location within $[\mathbf{l}_k^i, \mathbf{u}_k^i]$, obtained by minimizing the one-dimensional model $\hat{f}(c)$ fitted to the triplet $(c_j^{m+1}, c_j^i, c_j^{m+2})$.
	
	Then, the hyper-rectangle $\bar{\mathcal{X}}_k^h$, whose center $\mathbf{c}^h$ is closest to the approximate minimizer $\hat{c}^{\rm min}_j$, is determined as
	\begin{equation}
		\label{eq:hPOH}
		\bar{\mathcal{X}}_k^h
		=
		\argmin_{h \in \{i, m+1, m+2\}}
		\left( \left| \hat{c}^{\rm min}_j - c_j^h \right|, f(\mathbf{c}^h) \right).
	\end{equation}
	If two hyper-rectangles are located at the same distance from the approximate minimizer, the tie is resolved according to their ordering in the candidate set, i.e., the hyper-rectangle with the smallest index is selected.
	The hyper-rectangle $\bar{\mathcal{X}}_k^h$ is further subdivided if
	\begin{equation}
		\label{eq:subdivCond}
		\hat{f}^{\rm min} < f_k^{\rm min} - \varepsilon_{\mathrm{imp}}
		\quad \text{and} \quad
		\bar{d}_k^i \ge \varepsilon_{\mathrm{size}}.
	\end{equation}
\end{definition}

Consequently, \Cref{def_dynPartRule} facilitates the subdivision of hyper-rectangles in cases where the estimated minima demonstrate an improvement over the current optimal solution, with $\varepsilon_{\rm imp}$ controlling the expected enhancement, and $\varepsilon_{\text{size}}>0$ is the minimum side-length threshold (to prevent excessive refinement).
Similarly to \eqref{eqn_potOptRect2}, condition \eqref{eq:subdivCond} is needed to stop the algorithm from wasting function evaluations by partitioning hyper-rectangles where we can only expect a negligible improvement.

The left and middle parts of~\Cref{fig_example1} in \Cref{ex12} illustrate how dynamic partitioning would perform on a simple quadratic function.
If the predicted $\hat{f}^{\rm min}$ is better than $f^{\rm min}_k$, \Cref{def_dynPartRule} will continue to subdivide the hyper-rectangle containing the predicted minima. 
We repeat this process while the surrogate predicts a sufficiently promising improvement and the triplet step size remains above the user-prescribed limit; otherwise, the dynamic partitioning is terminated.
\begin{example}
	\label{ex12}
	The following example~\Cref{fig_example1} shows the performance of the suggested partitioning scheme utilized on a basic shifted \textit{Sphere} function (left) and a linear function (middle) described in \eqref{eq_prob1}. 
	In this example, we aim to achieve a target function value within an absolute value of $10^{-8}$. 
	In both examples, the black points denote evaluated sample locations, whereas the blue points and dashed lines illustrate the approximation employed by the dynamic partitioning mechanism.
	The \textnormal{\xdtcgl} algorithm successfully identified the solution without necessitating additional global searches. 
	On the right side of the figure, we replicated the experiment performed in~\cite{Jones2021_review} using the same linear function \eqref{eq_prob1}. 
	As the dimension increases, the number of distinct measure hyper-rectangles increases~\cite{Stripinis2021_pDIRECT_GLce}, rendering the isolation of the solution with the required accuracy expensive, even for straightforward instances. 
	
	The left side of the figure demonstrates the improved convergence of the proposed partitioning methodology relative to its \textnormal{\direct}-type counterparts when applied to various dimensions of the linear function. 
	The \textnormal{\xdtcgl} algorithm can isolate the solution in the evaluated straightforward instances with the necessary precision while requiring a minimal number of hyper-rectangular partitions, whereas other algorithms exhibit a significant ``global drag'' issue.
	Starting from the one-dimensional case, the original \textnormal{\direct} and \textnormal{\idtcgl} algorithms generate approximately $60\%$ more hyper-rectangles than the proposed \textnormal{\xdtcgl} before reaching the convergence threshold. 
	This disparity grows rapidly with dimension. 
	For example, the excess reaches approximately $99.8\%$ for \textnormal{\direct} already in dimension five, indicating that an increasingly large fraction of the search effort is spent exploring hyper-rectangles that do not contribute to faster convergence. 
	In contrast, the more locally focused \textnormal{\directc} and \textnormal{\plorc} algorithms perform considerably better in this respect. 
	Nevertheless, a substantial amount of search effort remains unproductive, with the corresponding excess reaching approximately $70\%$ and $48\%$, respectively, for $n {\ge} 10$.
	\begin{figure}[ht!]
		\centering
		\includegraphics[width=1\linewidth]{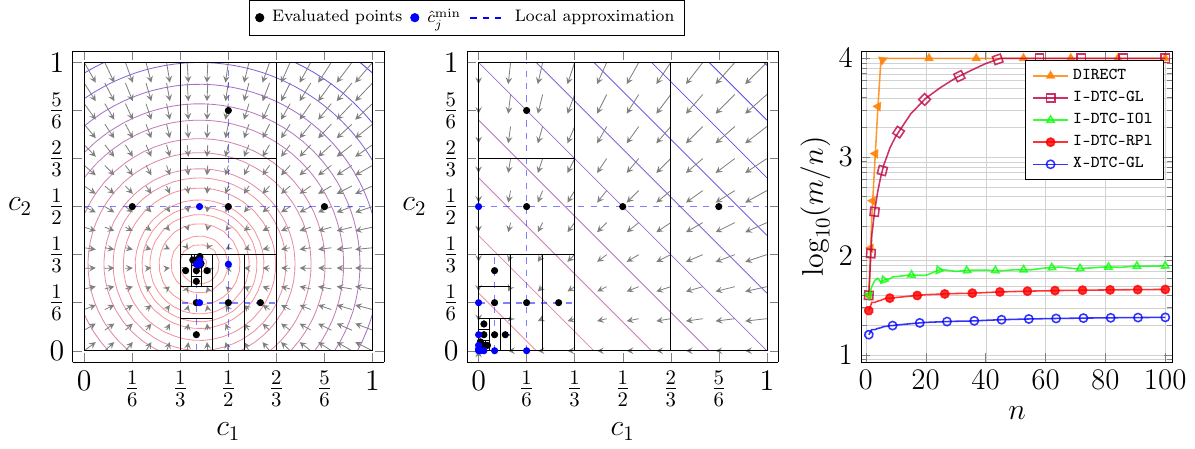}
		\caption{The left and middle panels illustrate the performance of the \xdtcgl{} algorithm on two straightforward instances: the two-dimensional shifted \textit{Sphere} function~\cite{DIRECTGOLib2023} and a linear function described in \eqref{eq_prob1}. The right panel compares the performance of \xdtcgl{} with relevant \direct-type counterparts on the linear function \eqref{eq_prob1} with different dimensions (lower curves indicate better performance).}
		\label{fig_example1}
	\end{figure}
	
	While this example investigates straightforward instances where dynamic partitioning naturally directly leads to the optimum, in complex multi-modal cases, the local models may initially be highly imprecise due to the small number of points and vast intervals used to fit the model.
	However, as the hyper-rectangles are iteratively reduced in size, the samples become denser, and the approximation might evolve very accurately.
\end{example}

\subsubsection{Algorithmic Steps, Hybridization, and Comment on the Convergence}

\paragraph{Algorithm workflow}
Algorithm~\ref{alg_algorithm2} provides the final pseudo-code for the \xdtcgl{} algorithm.
The algorithm requires three inputs: the objective function ($f$), the optimization domain ($\mathcal{X}$), and algorithmic options defining parameters and stopping criteria, such as the function target value ($f^{\rm target}$), the maximum number of function evaluations ($\texttt{M}_{\rm max}$), or the maximum number of iterations ($\texttt{K}_{\rm max}$).

The initialization phase (lines~\ref{alg_init}--\ref{alg_loclacounter_init}) largely follows the original \direct{} procedure, except that a hill-climber is executed in line~\ref{alg_hillclimber_init} to refine the initial minimum.
After the initialization phase (lines~\ref{alg_iterbegin}--\ref{alg_iterend}), the algorithm enters the main loop, which continues until at least one of the stopping criteria is satisfied.
In line~\ref{alg_findpoh}, the algorithm determines the set of (POHs), which are subsequently partitioned dynamically in lines~\ref{alg_partbegin}--\ref{alg_partend}.
Depending on the selected variant, the surrogate model used in the dynamic partitioning step (line~\ref{alg_surrogate}) is either \texttt{quadratic} ($Q$, default) or \texttt{linear} $(L)$. 
In the remainder of this paper, the corresponding variants are denoted by \xdtcgl$_Q$ and \xdtcgl$_L$, respectively.
The depth of dynamic partitioning is controlled by $\varepsilon_{\rm size}$ and $\varepsilon_{\rm imp}$, which act as numerical safeguards.
By default, they are both set to $10^{-16}$, slightly above the machine precision limit.
Similar safeguards are present in several \direct{} implementations, although they are often used implicitly and not explicitly discussed.
Upon completion, the algorithms provide the solution point $(\mathbf{c}^{\min}_k)$, the objective function value found $(f^{\min}_k)$, and algorithmic performance measures, including the number of function evaluations $(m)$, the number of iterations $(k)$, and time ($t$).

\SetEndCharOfAlgoLine{}
\begin{algorithm}[ht!]
	\footnotesize
	\nonl\textbf{Input:} \\
	\nonl \quad $f$: Objective function\;
	\nonl \quad $\mathcal{X}$: Decision space\;
	\nonl \quad \textit{OPT}: structure with optimization options\;
	\nonl\textbf{Output:} \\
	\nonl \quad  $f^{\rm min}_k$, $\mathbf{c}^{\rm min}_k$: Optimal solution and its corresponding point\; 
	\nonl \quad $t$ (time), $k$ (iterations), $m$ (function evaluations): Performance measures\;
	\algrule
	\textbf{Initialization}: Normalize $\mathcal{X}$ to $\bar{\mathcal{X}}$. Evaluate $f$ at $\mathbf{c}^1$. Set $f^{\rm min}_1 \leftarrow f(\mathbf{c}^1)$ and $\mathbf{c}^{\rm min}_1 \leftarrow \mathbf{c}^1$ and initialize $t$, $k \leftarrow 1$, $m \leftarrow 1$, $\lambda \leftarrow 0$, $H_1 \leftarrow \emptyset$, and \textit{stopping criteria}\label{alg_init} \; 
	Run hill-climber starting from $\mathbf{c}^1$. \label{alg_hillclimber_init} \\
	Update $f^{\rm min}_1, \mathbf{c}^{\rm min}_1, m$ and set $H_k \leftarrow H_k \cup i$.  \label{alg_loclacounter_init}\\
	\While(\tcp*[f]{\scriptsize algorithm iterations}){$f^{\rm target} < f^{\rm min}_k$ {\rm \textbf{and}} $m < \texttt{M}_{\rm max}$ {\rm \textbf{and}} $k < \texttt{K}_{\rm max}$}{ \label{alg_iterbegin}
		\textit{Selection}: Identify the index set $\mathbb{S}_k \subseteq \mathbb{I}_k$ of POHs. \tcp*[f]{\scriptsize selection method~\cite{Stripinis2018_DIRECT_GL}}\\ \label{alg_findpoh}
		\ForEach(\tcp*[f]{\scriptsize loop through all selected POHs}){$i \in \mathbb{S}_k$}{\label{alg_partbegin}
			\Repeat(\tcp*[f]{\scriptsize dynamic partitioning}){eq. \eqref{eq:subdivCond} is satisfied}{\label{alg_dynamicbegin} 
				Find the maximum side length $j$ of $\bar{\mathcal{X}}^i_k$ which is least split in $\mathcal{P}_k$.\\
				\textit{Sample} $\mathbf{c}^i \pm \bar{d}^i_k\mathbf{e}_j$, evaluate $f$, and update $f^{\rm min}_k, \mathbf{c}^{\rm min}_k,$ and $m$.\\
				\textit{Subdivide} $\bar{\mathcal{X}}^i_k$ (trisect) and update $\mathcal{P}_k \leftarrow \mathcal{P}_k \cup \bar{\mathcal{X}}^{m-1}_k \cup \bar{\mathcal{X}}^{m}_k$.\\
				\textit{Construct surrogate model} $\hat{f}(c)$ and find $\hat{f}^{\rm min}, \hat{c}^{\rm min}$. \label{alg_surrogate}\\
				Update the POH index $i$ using \eqref{eq:hPOH}.\label{alg_updatepoh}\\
				\If(\tcp*[f]{\scriptsize hybridization}){eq. \eqref{eq:subdivCond} {\rm \textbf{and}} $\hat{c}^{\rm min} \in (\mathbf{l}_k^i, \mathbf{u}_k^i)$ {\rm \textbf{and}} $i \notin H_k$}{\label{alg_minbegin}
					Run hill-climber starting from $\mathbf{c}^i_k$. \label{alg_hillclimber} \\
					Update $f^{\rm min}_k, \mathbf{c}^{\rm min}_k, m$ and set $H_k \leftarrow H_k \cup i$.  \label{alg_loclacounter}\\
				}\label{alg_minend}
			}\label{alg_dynamicend}
		}\label{alg_partend}
		Set $k \leftarrow k + 1$ and update $t$.\\
	}\label{alg_iterend}
	\textbf{Return}: $f^{\rm min}_k, \mathbf{c}^{\rm min}_k$, and performance measures ($t$, $k$, $m$).
	\caption{Main steps of \xdtcgl{} algorithm}
	\label{alg_algorithm2}
\end{algorithm}

\paragraph{Incorporating the hill-climber}
In addition to the initialization step, the hill-climber may also be invoked during the main iterations. 
In lines~\ref{alg_minbegin}--\ref{alg_minend} of Algorithm~\ref{alg_algorithm2}, a hill-climber is initiated from the center of the hyper-rectangle if: the approximation suggests that this region could potentially offer a better value; the predicted minimizer lies within the hyper-rectangle; and no local search has previously been performed from this region.
All function evaluations performed during the hill-climber search are counted toward the global evaluation counter $m$.
After each local run, $m$ is updated accordingly (lines~\ref{alg_loclacounter} and \ref{alg_loclacounter_init}), ensuring that comparisons with other methods remain fair.
The suggested hybridization strategy serves as a middle ground between existing approaches. 
For instance, algorithms like \birmin{} and \directrev{} initiate local searches only when there is significant improvement in $f^{\rm min}_k$, whereas methods like \dirmin{} and \directt{} trigger local searches from every POH or sampled point. 
In contrast, our approach proactively initiates local searches without waiting for $f^{\min}_k$ to improve, while also avoiding the execution of local searches in regions deemed unattractive.

\paragraph{Comments on the convergence}
The convergence properties of \direct-type algorithms have been extensively reviewed and investigated in the literature, as can be seen in relevant references~\cite{ Jones1993_DIRECT, Finkel2006_DIRECT_m, Stripinis2024_HALRECT, Paulavicius2018_BIRECT, Sergeyev2006_ADC}. 
These algorithms typically exhibit a type of convergence known as ``everywhere-dense''.
Convergence can be ensured by only assuming continuity of the objective function, at least in the vicinity of global minima.
Since the selection scheme used in the \xdtcgl{} always includes at least one hyper-rectangle from the group of hyper-rectangles with the largest measure $\delta_k^{\rm max}$ in the set of POHs, the \xdtcgl{} convergence can be proven using the same reasoning as for other \direct-type algorithms.

\paragraph{Comments on the implementation}
In practical black-box scenarios, the objective function may not be evaluable at all sampled locations due to simulation failures, infeasible configurations, or data sparsity. 
The proposed \xdtcgl{} framework operates on point-wise evaluations and does not require an analytic representation of the objective; however, it assumes that each queried point returns a finite scalar value that can be evaluated relative to other candidates.
When evaluations are unavailable at specific locations, standard handling strategies such as rejection, penalization~\cite{Stripinis2019_DIRECT_GLce, Stripinis2021_DIRECT_GLh}, or surrogate-based interpolation from available data should additionally be employed.
These mechanisms would allow the algorithm to remain applicable in settings where the objective is defined only on a subset of the feasible region.

\subsection{Component Analysis, Robustness Assessment, and Limitations}
\label{sec_ablt}
This section evaluates the contribution of the proposed individual mechanisms, investigates the algorithm’s sensitivity to problem perturbations, assesses its performance across different problem types, and identifies potential limitations.
For this purpose, we employ the BBOB benchmark suite~\cite{BBOB_Hansen2009} available through the \texttt{IOHprofiler}~\cite{IOHprofiler} Python interface. 
To ensure a robust evaluation, we considered five dimensions $n {\in} \{2,3,5,10,20\}$ and the first five instances of each BBOB function, resulting in $600$ total runs per algorithm.
All algorithms were evaluated using a budget of $n {\times} 10^5$ function evaluations, with the goal of achieving an absolute error below $10^{-4}$. 
The experiments were conducted in \texttt{MATLAB R2023a} on a system running Microsoft Windows 10, equipped with an 8th-generation Intel Core i7-8750H processor (6 cores) and 16 GB of RAM.

For the ablation study, we derive several algorithmic variants from the baseline \idtcgl. 
Specifically, we consider the baseline \idtcgl{} and eight extensions obtained by adding individual components, as summarized in \Cref{tab_ablation_components}. 
Each extension is studied in two versions, depending on the surrogate model used in the dynamic partitioning step: a quadratic version $(Q)$ and a linear version $(L)$. 
Accordingly, we consider: (i) \texttt{DP}$_{Q}$ and \texttt{DP}$_{L}$, which augment the baseline with dynamic partitioning; 
(ii) \texttt{1LS-DP}$_{Q}$ and \texttt{1LS-DP}$_{L}$, which further add a single local solver run at initialization; (iii) \texttt{LS}$_{Q}$ and \texttt{LS}$_{L}$, which incorporate the proposed local solver usage strategy without dynamic partitioning; and (iv) the full method, \xdtcgl$_Q$ and \xdtcgl$_L$, which combine all components.

\begin{table}[t]
	\centering
	\small
	\caption{Algorithmic components activated in the investigated variants. For each component, the corresponding lines in Algorithm~\ref{alg_algorithm2} are indicated in the second header row. A plus sign ($+$) denotes that the corresponding steps are active in a given variant, whereas a minus sign ($-$) denotes that they are inactive.}
	\label{tab_ablation_components}
	\begin{tabularx}{\textwidth}{lXXX}
		\toprule
		\multirow{2}{*}{Algorithm} 
		& Dynamic partitioning 
		& Init.\ local solver 
		& LS strategy \\
		& Alg.~\ref{alg_algorithm2}, lines~\ref{alg_dynamicbegin}, \ref{alg_dynamicend}, \ref{alg_surrogate}--\ref{alg_updatepoh}
		& Alg.~\ref{alg_algorithm2}, lines~\ref{alg_hillclimber_init}--\ref{alg_loclacounter_init}
		& Alg.~\ref{alg_algorithm2}, lines~\ref{alg_minbegin}--\ref{alg_minend} \\
		\midrule
		\idtcgl                                   & $-$ & $-$ & $-$ \\
		\texttt{DP}$_Q$ / \texttt{DP}$_L$         & $+$ & $-$ & $-$ \\
		\texttt{1LS-DP}$_Q$ / \texttt{1LS-DP}$_L$ & $+$ & $+$ & $-$ \\
		\texttt{LS}$_Q$ / \texttt{LS}$_L$         & $-$ & $+$ & $+$ \\
		\xdtcgl$_Q$ / \xdtcgl$_L$                 & $+$ & $+$ & $+$ \\
		\bottomrule
	\end{tabularx}
\end{table}

\subsubsection{Performance Across Different Problem Types}
\label{sec_problemTypes}

\begin{figure}[ht!]
	\centering
	\includegraphics[width=1\linewidth]{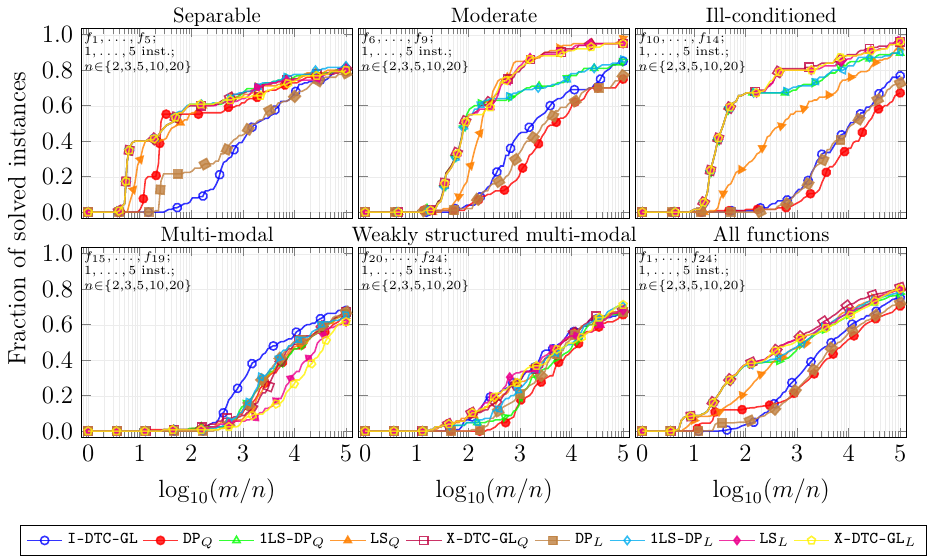}
	\caption{Data profiles showing the proportion of problem instances solved (higher is better) as a function of the allowable number of function evaluations across five BBOB problem classes and overall.}
\label{figs_DatProf21}
\end{figure}

The performance of all nine algorithmic variants is illustrated in \Cref{figs_DatProf21}. 
Each panel of the figure corresponds to a different BBOB problem class and aggregates results across all instances and problem dimensions, while the final panel reports the overall performance aggregated over all problem classes.

A first observation is that dynamic partitioning alone (both variants \texttt{DP}$_{Q}$ and \texttt{DP}$_{L}$) provides only limited benefits and may even degrade the performance of the original baseline \idtcgl{} algorithm. 
This behavior can be explained by the fact that the algorithm initially lacks a sufficiently good estimate of $f^{\min}$. 
Consequently, the dynamic partitioning mechanism may prioritize subdivisions that appear promising but do not lead to meaningful improvements. 
As a result, the algorithm may spend a considerable number of function evaluations exploring subregions before identifying a sufficiently strong minima value.
However, when the local solver is executed once at initialization (\texttt{1LS-DP}$_{Q}$/\texttt{1LS-DP}$_{L}$), the performance improves substantially for most problem classes. 
The initial local search provides a better estimate of $f^{\min}$, which in turn allows the dynamic partitioning mechanism to guide the search more effectively. 
Although these variants still do not match the performance of the full \xdtcgl$_{Q}$ and \xdtcgl$_{L}$ algorithms for moderately and ill-conditioned problems, they consistently outperform the baseline \idtcgl{} on most problem types.

The variants employing only the local solver usage strategy (\texttt{LS}$_{Q}$ / \texttt{LS}$_{L}$) are generally competitive with the full \xdtcgl$_{Q}$ and \xdtcgl$_{L}$ methods. 
Nevertheless, \texttt{LS}$_{Q}$ performs noticeably worse than \xdtcgl$_{Q}$ and \xdtcgl$_{L}$   on ill-conditioned problems, while \texttt{LS}$_{L}$ ranks among the least competitive methods on multi-modal problems.
Overall, \xdtcgl$_{Q}$ achieves the best performance in most cases, solving the largest proportion of problem instances across the considered problem classes.

The introduced modifications do not always provide consistent improvements across all problem types. 
For highly challenging, weakly structured multi-modal problems, the performance of most variants is similar to that of the baseline \idtcgl{}. 
In particular, all algorithms solve approximately $65.6\%-70.4\%$ of the instances, with the \texttt{1LS-DP}$_{L}$ achieving the highest success rate. 
A notable limitation is observed for multi-modal problems, where the baseline solves the largest proportion of instances (about $68\%$), and its curve remains consistently above the other methods, while the second-best method, \xdtcgl$_{Q}$, solves approximately $67.2\%$.

\subsubsection{Sensitivity to Instance Perturbations}

To evaluate the algorithms' sensitivity to changes in the optimization domain and assess whether the proposed modifications affect robustness, we analyzed their performance across the five randomized BBOB instances for each problem. 
These instances introduce shifts and rotations of the search space, along with changes in the objective function minima values, providing a natural mechanism for evaluating robustness to such perturbations.

\begin{figure}[ht!]
	\centering
	\includegraphics[width=1\linewidth]{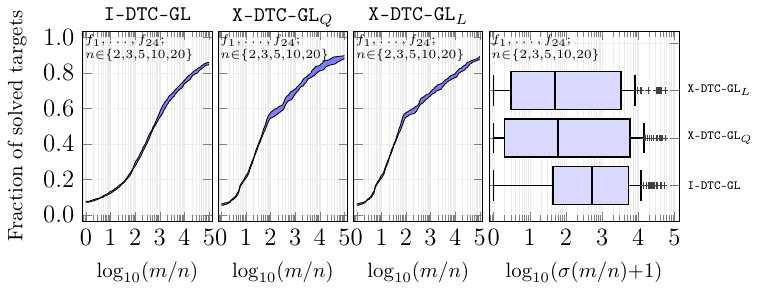}
	\caption{The three plots on the left show ECDFs of the fraction of precision targets reached (higher is better); the filled region indicates the min--max range across the five instances (narrower is better). The plot on the right depicts boxplots of the standard deviation of the number of function evaluations required to reach the target across instances, aggregated over all problems (lower is better).}
	\label{figs_robustness}
\end{figure}

\Cref{figs_robustness} summarizes the variability across instances. 
The left three panels (each for a different algorithm) show empirical cumulative distribution functions (ECDFs) of the fraction of precision targets achieved over the evaluation budget, with shaded regions indicating the min–max range across the five instances. 
The targets were defined using $51$ absolute precision levels logarithmically spaced in $10^{[-4,2]}$, similar to the setup used in the COCO benchmarking platform~\cite{hansen2021coco}.
The shaded regions remain narrow for all three algorithms, indicating minimal variability across instances. 
This suggests that both the baseline algorithm (\idtcgl) and the proposed modifications (\xdtcgl$_Q$ and \xdtcgl$_L$) are robust and largely insensitive to the introduced perturbations, consistently approaching the desired target precision using a similar number of function evaluations.

The right panel of \Cref{figs_robustness} presents boxplots of the standard deviation of the number of function evaluations required to reach the prescribed target value across the five instances of each problem, aggregated over all problems.
For both developed algorithms, the median standard deviation is approximately $50{\times}n$ function evaluations. 
In contrast, for the original algorithm (\idtcgl), it is about $550{\times}n$, indicating substantially higher variability across instances.
However, the third quartile and higher values are comparable across all three algorithms.

The few extreme outliers correspond to problems in which some instances require substantially more evaluations than others. 
Since these outliers appear for all algorithms, they likely reflect inherent variability in the difficulty of specific benchmark instances rather than algorithm instability.

\subsubsection{Runtime Overhead Assessment}

We additionally report wall-clock execution times ($t$) to quantify the computational overhead introduced by hybridization, surrogate construction, and the adaptive decision mechanism. 
For this purpose, we compare three algorithms: the original baseline method \idtcgl{} and its two modified variants, \xdtcgl$_Q$ and \xdtcgl$_L$. 

\begin{figure}[ht!]
	\centering
	\includegraphics[width=1\linewidth]{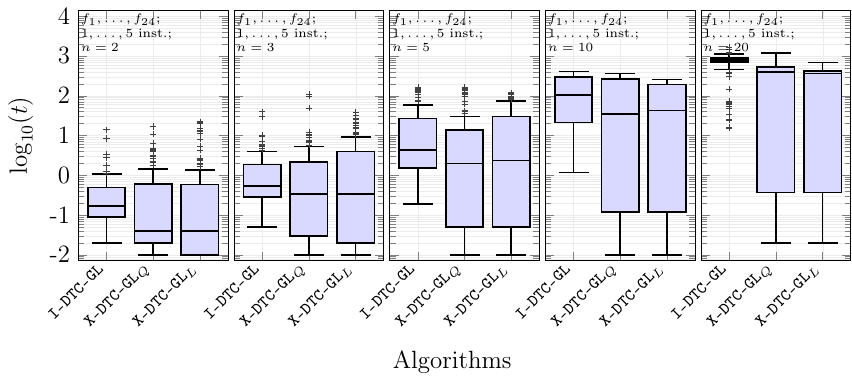}
	\caption{Boxplots of execution time $t$ (lower is better) for three algorithms on the $24$ BBOB benchmark functions, evaluated over five instances. Each subplot corresponds to a different problem dimension $n \in \{2,3,5,10,20\}$.}
	\label{figs_time}
\end{figure}

\Cref{figs_time} shows the distribution of total execution time over the $24$ BBOB functions and five instances for all considered dimensions. 
The reported time corresponds to the duration required to reach the prescribed absolute error threshold; in case of failure, it reflects the time needed to exhaust the entire evaluation budget during the search.

The minimum, first quartile, and median values of both modified variants are consistently lower than those of the original algorithm across all five panels. 
This follows from the fact that the modifications reduce the number of function evaluations on most problems (as we previously showed); consequently, the overall execution time decreases accordingly.

More informative are the third quartile, the maximum values, and the upper outliers, which are comparable across all panels and algorithms. These statistics largely correspond to runs in which the complete evaluation budget is exhausted. 
At this level, the total runtime reflects the cost of completing the entire optimization process rather than the cost of early convergence.
Importantly, the additional computations introduced do not lead to a noticeable increase in wall-clock time. 
In dimensions $n \geq 10$, the modified algorithms are even slightly faster. 
This effect can be attributed to the hybrid structure: local subroutines, inexpensive surrogate-based steps, and dynamic partitioning may consume a portion of the evaluation budget at relatively low computational cost, thereby reducing the number of iterations the core optimization routine can perform. 
As a result, the overall runtime remains comparable or occasionally lower, despite the added algorithmic components.

\subsubsection{Limitations of the Proposed Algorithm}

The proposed extensions improve performance on many problem classes; however, as demonstrated in~\Cref{sec_problemTypes}, their benefits are not universal. 
The effectiveness of the hybridization strategy and dynamic partitioning depends on specific properties of the objective function.
In the following paragraphs, we describe two limitations identified during the study.

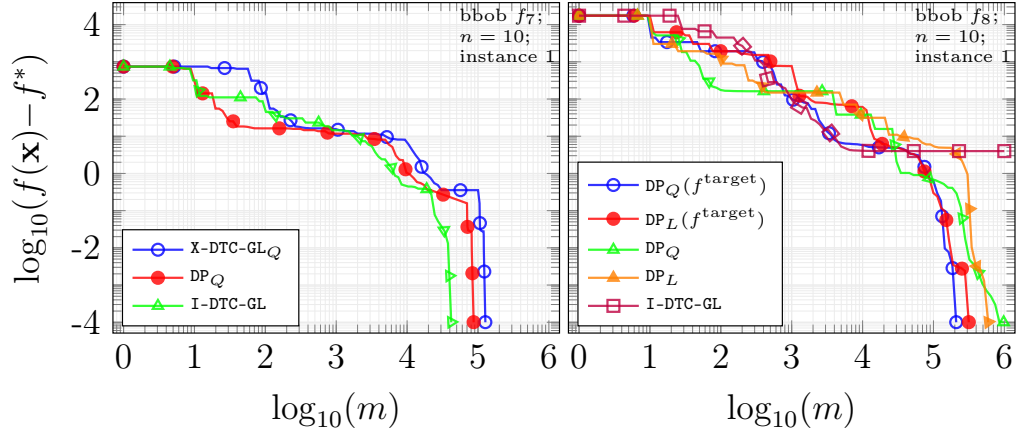
\begin{figure}[ht!]
	\resizebox{\textwidth}{!}{
		\begin{tikzpicture}
			\begin{groupplot}[group style={group size=2 by 1, y descriptions at=edge left, horizontal sep=3pt,group name = Lib,}, ymin=1e-4, ymax=20000, ytick distance=10, ymode=log, enlarge y limits=0.035, xmin=1, xmax=1000000, xtick distance=10, xmode=log, enlarge x limits=0.025, grid=both, grid style={line width=.1pt, draw=gray!15}, yticklabels={,-4,,-2,,0,,2,,4,}, height=0.4\textwidth, xticklabels={0,0,1,2,3,4,5,6}, ylabel={$ \log_{10}(f(\mathbf{x}){-}f^\ast) $}, xlabel={$\log_{10}(m)$}, legend style={at={(0.02,0.02)},anchor=south west,font=\tiny}, legend cell align={left}, width=0.5\textwidth] 
				\nextgroupplot
				\pgfplotsinvokeforeach{1,...,3}{\addplot[s_a#1] table[x=T#1,y=x#1] {conv.txt};}
				\legend{\xdtcgl$_Q$,\texttt{DP}$_{Q}$,\idtcgl}
				\nextgroupplot  
				\pgfplotsinvokeforeach{1,...,5}{\addplot[s_a#1] table[x=G#1,y=y#1] {conv.txt};}
				\legend{\texttt{DP}$_{Q}$($f^{\rm target}$),\texttt{DP}$_{L}$($f^{\rm target}$),\texttt{DP}$_{Q}$,\texttt{DP}$_{L}$,\idtcgl} 
				
			\end{groupplot}
			\node[draw=none, fill=none, anchor=north east, inner sep=1pt] at (Lib c1r1.north east)
			{\tiny  \begin{tabular}{@{}l@{}}
					bbob $f_{7}$; \\ $n=10$; \\ instance $1$
			\end{tabular}};
			\node[draw=none, fill=none, anchor=north east, inner sep=1pt] at (Lib c2r1.north east)
			{\tiny  \begin{tabular}{@{}l@{}}
					bbob $f_{8}$; \\ $n=10$; \\ instance $1$
			\end{tabular}};
	\end{tikzpicture}}
	\caption{Convergence plots for the \textit{Step Ellipsoidal} (left) and \textit{Rosenbrock} (right) test functions in dimension $10$.}
	\label{figs_example3}
\end{figure}

\paragraph{Hybridization and dynamic partitioning overhead}
It is well known that local solvers do not always perform well~\cite{conn2009derivativefree}. 
Therefore, in the proposed framework, they may introduce additional computational overhead without improving the solution quality. 
This may occur with non-smooth functions or in regions with flatness, where gradient information is difficult to approximate efficiently.
The left panel of~\Cref{figs_example3} and \Cref{ex13} illustrates and explains that hybridization and dynamic partitioning may negatively affect performance.

\begin{example}
	\label{ex13}
	The left panel of \Cref{figs_example3} shows the convergence plots of three algorithms for the first instance of the \textit{Step Ellipsoidal} test function in dimension $10$. 
	The convergence curve of the \xdtcgl$_Q$ algorithm is the slowest. 
	This behavior occurs because dynamic partitioning identifies promising regions and triggers the hill-climber frequently ($1902$ times). 
	However, the \textit{Step Ellipsoidal} function consists of many plateaus of different sizes, and the gradient is zero almost everywhere except in a small region near the global optimum.
	As a result, many local-search calls are performed, consuming a substantial number of function evaluations ($40289$, approximately $21$ per call), while none of them produce an improvement.
	
	Dynamic partitioning alone (\texttt{DP}$_Q$), without the local solver, converges faster than \xdtcgl$_Q$. 
	However, the baseline \idtcgl{} eventually achieves the fastest convergence. 
	Although the \textit{Step Ellipsoidal} function globally retains a quadratic ellipsoidal structure, consequently, the \texttt{DP}$_Q$ variant initially approaches the solution faster because the surrogate model can still capture the global quadratic trend of the landscape.
	However, once the search enters plateau regions, the surrogate approximation becomes less informative, as many sampled points share identical function values. 
	In such regions, the model cannot reliably predict promising directions, which slows further progress.
\end{example}

\paragraph{Limitations of the surrogate approximation}
As shown in \Cref{sec_problemTypes}, dynamic partitioning alone may be ineffective and can even degrade the performance of the baseline \idtcgl{} algorithm. 
For dynamic partitioning to work effectively, a reasonably accurate estimate of the minimum value is needed. 
Otherwise, if the initial minimum is poor, many selected potentially optimal hyper-rectangles might appear promising, leading the algorithm to partition them excessively, even if they do not contain true minima. 
To mitigate this issue, a local solver is executed during initialization (Algorithm~\ref{alg_algorithm2}, line~\ref{alg_hillclimber_init}) to refine the initial minimum before entering the main iteration loop.
\Cref{ex13} and the right panel of \Cref{figs_example3} illustrate and explain that providing a better initial value improves the performance of the dynamic partitioning strategy.

\begin{example}
	\label{ex14}
	To illustrate this limitation, we consider the baseline \idtcgl{}, the two dynamic partitioning variants \texttt{DP}$_{Q}$ and \texttt{DP}$_{L}$, and modified versions where the target value ($f^{\rm target}$) is used instead of $f^{\rm min}_k$ in the condition of \Cref{eq:subdivCond}.
	The right panel of \Cref{figs_example3} shows the convergence plots of these variants for the $10$-dimensional Rosenbrock function. 
	In higher dimensions, this problem contains a local optimum with a large attraction basin, which causes the baseline \idtcgl{} to fail to reach the global solution.
	Both dynamic partitioning variants (\texttt{DP}$_{Q}$ and \texttt{DP}$_{L}$) can solve the problem, but they require a large number of function evaluations to reach the target.
	When the target value ($f^{\rm target}$) is used to control when dynamic partitioning is applied, the performance of both variants improves significantly.
	In particular, \texttt{DP}$_{Q}$($f^{\rm target}$) reaches the desired solution using approximately three times fewer function evaluations, while the \texttt{DP}$_{L}$($f^{\rm target})$ variant requires roughly twice fewer evaluations.
	This experiment suggests that providing a reliable estimate of the minimum can substantially improve the effectiveness of dynamic partitioning. 
	Note, however, that in practice the value $f^*$ is not known and is used here only for experimental analysis.
\end{example}

\section{Experimental Results and Discussion}
\label{sec_experiments}

\subsection{Experimental settings}

\paragraph{Algorithms and experimental setup}
The performance of the two versions of the \xdtcgl{} algorithm (with linear -- \xdtcgl$_L$ and quadratic -- \xdtcgl$_Q$ approximations) was evaluated compared to six \direct-type algorithms and the Naive Multi-scale Search Optimization (\nmso) algorithm~\cite{Abdullah2016_NMSO}, which utilizes a partitioning approach similar to that of \direct, and showed excellent performance in GECCO'15 BBO competition.
Among the \direct-type, we included \idtcgl{} as this algorithm forms the basis for the suggested \xdtcgl. 
Furthermore, the evaluation included the emerging \directt{} algorithm and two highly effective and widely adopted algorithms, \dirmin{} and \birmin.
Furthermore, the analysis encompassed two canonical algorithms, the \direct{} and \directc.
The main benchmark experiments were conducted under the same computational setup as the ablation study (\Cref{sec_ablt}). 
Each algorithm was allowed a budget of $n{\times}10^5$ function evalutions and success was declared when the absolute error fell below $10^{-4}$.

To better determine the impact of the \direct{} algorithm in hybrid approaches, in this study we used the same hill-climbing algorithm for all hybrid approaches. 
The local refinement stage was implemented using \texttt{MATLAB}'s \texttt{fmincon} function with the \texttt{sqp} algorithm option. 
Bound constraints were handled directly through the lower and upper bound arguments of \texttt{fmincon}. 
The SQP method was applied with its default configuration~\cite{MATLAB2023}, with the exception that the maximum evaluation budget per run was established at $n{\times}10^3$ for all algorithms, except for the \directt{}, for which $n{\times}10^2$ was recommended in \cite{Kanayama2023_tDIRECT}. 
Function evaluations performed by the SQP procedure were included in the total evaluation counter of the corresponding \direct-based algorithm.


\paragraph{Benchmark problems}
The recently extended \directgolib{} benchmark library~\cite{DIRECTGOLib2023} was used as the basis for testing the considered algorithms. 
The library encompasses an extensive collection of functions sourced from diverse origins, encompassing classical, emerging (ABS~\cite{ABS_Kudela2022} and Layeb~\cite{Layeb2022}), and widely-utilized (BBOB~\cite{BBOB_Hansen2009} and CEC~\cite{liang2013problem, wu2017problem}) functions. 
These functions exhibit a range of characteristics concerning differentiability, separability, multi-modality, and flatness.

We followed the settings and transformations of a recent study~\cite{Stripinis2024_benchmark} that utilized $324$ box-constrained test functions, investigating scalable ones with dimensions $2$, $5$, $10$, and $20$.
Each function was evaluated in five different instances with distinct random domain shifts, and the last three instances additionally involved random rotations, resulting in a total of $4035$ instances. 
All randomizations were generated using fixed seeds to ensure reproducibility.
These transformations also prevent the optimum from coinciding with the origin, ensuring that none of the algorithms can locate the solution through initial sampling alone, which is common in many artificial benchmark problems.

We selected four balanced and representative benchmark suites, each of size $50$ instances, using different instance selection methods~\cite{Stripinis2024ISM, Cenikj2023, Dietrich2024Impact}. 
An overview of the four suites of selected instances, together with brief summaries of their respective selection methodologies, is provided in \Cref{tab_ISA}.
The last column of the table also presents the boxplots of the success rates achieved using the $26$ algorithms from the open-sourced data of the recent benchmarking study~\cite{Stripinis2024_benchmark}. 
This information provides a better understanding of the complexity of each selected benchmark suite.
This benchmarking method was chosen to address the challenges associated with analyzing data from inadequately structured benchmark suites, as discussed in~\cite{Stripinis2024ISM}.
\begin{table}[ht!]
	\scriptsize
	\caption{Benchmark suites employed for experimental evaluation of the algorithms.}
	\begin{tabular*}{\textwidth}{@{\extracolsep{\fill}}cclp{3.25cm}cc}
		\toprule
		Benchmark & \multicolumn{3}{c}{Instance selection method} & \# of & Complexity of the \\
		\cmidrule{2-4}
		Suite & Ref. & Space & Description & inst. & benchmark suites$^c$ \\
		\midrule
		\textit{BS1-50} & \cite{Cenikj2023} & ELA$^a$ & Clusters instances using cosine similarity  & $50$ &
		\raisebox{-0.6\totalheight}{
			\begin{tikzpicture}
				\begin{axis}[height=0.15\textwidth, width=0.3\textwidth, xmin=0, xmax=1, xtick distance=0.2,yticklabel=\empty, tick style={draw=none}, xticklabel style={/pgf/number format/.cd, fixed, fixed zerofill, precision=1,}, tick label style={font=\tiny}, boxplot/every median/.style={red, thick, line width=0.5pt}, boxplot/every whisker/.style={black, ultra thick, densely dotted, line width=0.5pt}]
					\addplot+[boxplot prepared={median=0.64, upper quartile=0.72, lower quartile=0.565, upper whisker=0.82, lower whisker=0.14},] coordinates {};
				\end{axis}
		\end{tikzpicture}} \\
		\midrule
		\textit{BS2-50} & \cite{Dietrich2024Impact} & ELA$^a$ & Maximize diversity using Manhattan distance & $50$ & 
		\raisebox{-0.6\totalheight}{
			\begin{tikzpicture}
				\begin{axis}[height=0.15\textwidth, width=0.3\textwidth, xmin=0, xmax=1, xtick distance=0.2,yticklabel=\empty, tick style={draw=none}, xticklabel style={/pgf/number format/.cd, fixed, fixed zerofill, precision=1,}, tick label style={font=\tiny}, boxplot/every median/.style={red, thick, line width=0.5pt}, boxplot/every whisker/.style={black, ultra thick, densely dotted, line width=0.5pt}]
					\addplot+[boxplot prepared={median=0.74, upper quartile=0.795, lower quartile=0.625, upper whisker=0.84, lower whisker=0.3},] coordinates {};
				\end{axis}
		\end{tikzpicture}} \\
		\midrule
		\textit{BS3-50} & \cite{Stripinis2024ISM} & AP$^b$ & Maximize diversity of algorithms' runtime & $50$ & 
		\raisebox{-0.6\totalheight}{
			\begin{tikzpicture}
				\begin{axis}[height=0.15\textwidth, width=0.3\textwidth, xmin=0, xmax=1, xtick distance=0.2,yticklabel=\empty, tick style={draw=none}, xticklabel style={/pgf/number format/.cd, fixed, fixed zerofill, precision=1,}, tick label style={font=\tiny}, boxplot/every median/.style={red, thick, line width=0.5pt}, boxplot/every whisker/.style={black, ultra thick, densely dotted, line width=0.5pt}]
					\addplot+[boxplot prepared={median=0.38, upper quartile=0.6, lower quartile=0.28, upper whisker=0.76, lower whisker=0},] coordinates {};
				\end{axis}
		\end{tikzpicture}} \\
		\midrule
		\textit{BS4-50} & \cite{Stripinis2024ISM} & AP$^b$ \& ELA$^a$ & Maximizes diversity using statistical tests and Euclidean distance & $50$ &
		\raisebox{-0.6\totalheight}{
			\begin{tikzpicture}
				\begin{axis}[height=0.15\textwidth, width=0.3\textwidth, xmin=0, xmax=1, xtick distance=0.2,yticklabel=\empty, tick style={draw=none}, xticklabel style={/pgf/number format/.cd, fixed, fixed zerofill, precision=1,}, tick label style={font=\tiny}, boxplot/every median/.style={red, thick, line width=0.5pt}, boxplot/every whisker/.style={black, ultra thick, densely dotted, line width=0.5pt}]
					\addplot+[boxplot prepared={median=0.36, upper quartile=0.475, lower quartile=0.285, upper whisker=0.58, lower whisker=0.12},] coordinates {};
				\end{axis}
		\end{tikzpicture}} \\
		\bottomrule
	\end{tabular*}
	\scriptsize{$^a$ Exploratory landscape analysis~\cite{Mersmann2011ELA}, $^b$ Algorithm performance, $^c$ Success rates on the selected instances using $26$ algorithms from~\cite{Stripinis2024_benchmark}}\\
	\label{tab_ISA}
\end{table}

\subsection{Solution Discovery Efficiency}

In this section, we comprehensively analyze the performance efficiency of the algorithms employed to solve four distinct benchmark suites, utilizing both the performance~\cite{Dolan2002_performance_profiles} and data~\cite{More2009_data_profiles} profiles.
Both of these data evaluation tools only consider instances in which the algorithms delivered the solutions with the required accuracy.

\begin{figure}[ht!]
	\centering
	\includegraphics[width=1\linewidth]{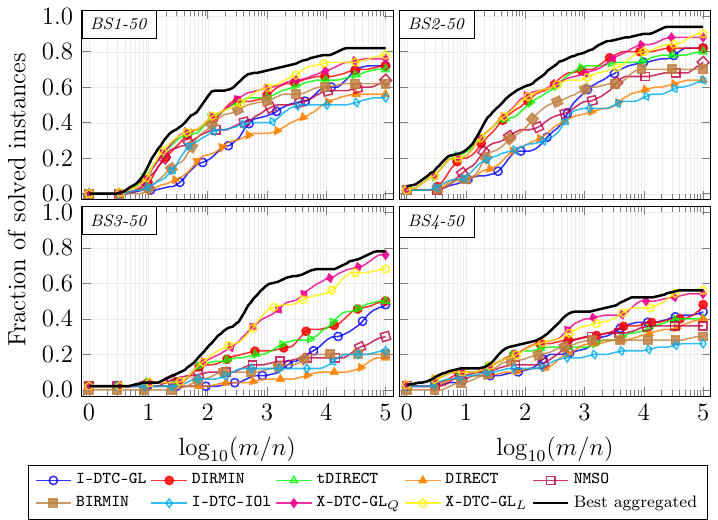}
	\caption{The data profiles illustrate the proportion of instances successfully solved (higher is better) as a function of the allowable number of function evaluations across four benchmark suites.}
	\label{figs_DatProf}
\end{figure}

The data profiles for the four benchmark suites are illustrated in \Cref{figs_DatProf}. 
The horizontal axis represents the number of function evaluations, whereas the vertical axis denotes the proportion of solved instances.
Additionally, the black curve in each subplot outlines the best aggregated performance achieved by any of the algorithms employed in the comparative study. 
This visualization clarifies the upper bound of performance attainable by the algorithms under consideration.

The analysis demonstrates that the final success rates of the algorithms in the first two benchmark suites (\textit{BS1} and \textit{BS2}) are more significant than in the remaining two benchmark suites. 
However, when comparing the overall ranking of the algorithms in all four subplots, the results are almost identical, with some minor differences. 
In all subplots, the performance of both versions of the \xdtcgl{} algorithm consistently stands out, solving a higher number of instances when compared to any other algorithm. 
Nevertheless, the best aggregated black curve suggests that either version of the \xdtcgl{} algorithm does not consistently address instances that other algorithms can solve. 
Specifically, for \textit{BS1}, \textit{BS2}, and \textit{BS4}, the \xdtcgl$_Q$ algorithm failed to solve one to two such instances, while in the \textit{BS3} benchmark suite, other competing algorithms successfully addressed six instances that remained unsolved by the \xdtcgl$_Q$ algorithm.
Despite this, the effectiveness of both versions of the \xdtcgl{} algorithm is most notable in the \textit{BS3} benchmark suite.

While data profiles show a general view of solved instances, performance profiles are specifically designed to compare the performance of specific algorithms to a set of algorithms.
Performance profiles assess the overall performance of algorithms using a performance ratio $(r_{i,a})$. 
This ratio is computed as:
\begin{equation}
	r_{i,a} = \frac{t_{i,a}}{\min\{ t_{i,a} : a \in \mathscr{A} \}},
\end{equation}
where $t_{i,a} > 0$ is the metric for algorithm $a$ solving instance $i$, and $\min\{t_{i,a}: a \in \mathscr{A} \}$ is the best metric for the instance. 
The performance profile $(\rho_a(\lambda))$ of an algorithm $a$ is derived from the cumulative distribution function of the performance ratio:
\begin{equation}
	\label{eq:perprof}
	\rho_a(\lambda) = \frac{1}{|\mathscr{I}|} |\{ i \in \mathscr{I} : r_{i,a} \le \lambda \}|, \quad \lambda \ge 1,
\end{equation}
where $|\mathscr{I}|$ is the number of problems. 
$\rho_a(\lambda)$ represents the probability that $r_{i,a}$ for each $i \in \mathscr{I}$ is within a factor $\lambda$ of the best ratio. 

\begin{figure}[ht!]
	\centering
	\includegraphics[width=1\linewidth]{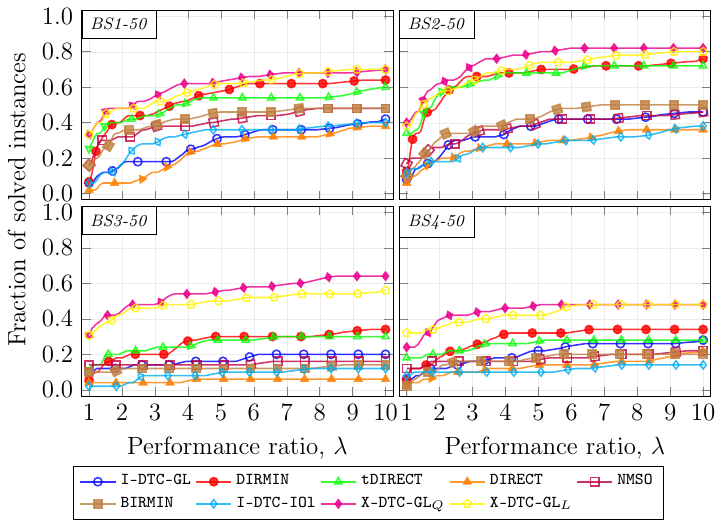}
	\caption{An analysis of the function evaluation metrics for the considered algorithms, utilizing four benchmark suites and employing performance profiles within the interval $\lambda \in [1,10]$ (higher curve indicates better performance).}
	\label{figs_PerProf}
\end{figure}

Performance profiles in \Cref{figs_PerProf} enable a comparison of algorithm performance across multiple instances in $\mathscr{I}$. 
A higher $\rho_a(\lambda)$ indicates better performance, and $\rho_a(1)$ indicates the fraction of instances where algorithm $a$ achieves the best performance.
Based on the performance profiles, it is evident that the developed algorithm demonstrates superior outcomes in three out of four benchmark suites. 
It exhibits the highest efficiency by achieving the most wins and solving the most instances with the smallest number of function evaluations. 
However, with regard to the number of wins, most algorithms also yield a reasonable number of wins. 
Notably, the \directt{} algorithm outperforms both versions of the proposed \xdtcgl{} algorithm in the second benchmark suite, most efficiently solving one-third of the instances.

The performance curves of both versions of the \xdtcgl{} algorithm consistently surpass those of other algorithms across three subplots, indicating its superior performance relative to the comparison algorithms. 
Notably, for the \textit{BS2} benchmark suite, the \xdtcgl$_Q$ and \xdtcgl$_L$ algorithms required a performance ratio of $\lambda$ to $1.2$ to position its curve above those of the other algorithms.
When the performance ratio reaches ten, the \xdtcgl$_Q$ algorithm provides solutions within ten times the number of function evaluations compared to the best-performing algorithms in approximately $72.5\%$, $75\%$, $47.5\%$ and $50\%$ of the instances for each of the four benchmark suites.

\begin{figure}[ht!]
	\begin{minipage}[t]{0.4\linewidth}
		\centering
		\begin{tikzpicture}[scale=0.5]
			\def\myalgslist{\idtcgl, \dirmin, \directt, \direct, \nmso, \birmin, \directc, \xdtcgl$_Q$, \xdtcgl$_L$}
			\pgfplotstableread{frt1.txt}\datatable
			\foreach \j in {0,...,8}{
				\foreach \i in {0,...,8}{
					\pgfplotstablegetelem{\j}{[index]\i}\of{\datatable}
					\let\colorvalue\pgfplotsretval
					\node[fill=blue!\colorvalue!white, draw=black, minimum size=5mm, line width=0.01pt] at (\i,\j) {};
				}
			}
			\foreach \i [count=\xi from 0] in \myalgslist{
				\node[rotate=90, anchor=east] at (\xi,-0.5) {\scriptsize \i};
			}
			\foreach \i [count=\xi from 0] in \myalgslist{
				\node[anchor=east] at (-0.5, \xi) {\scriptsize \i};
			}
			\draw[black, very thick, line width=0.5mm] (-0.6,-0.6) rectangle (8.6,8.6);
		\end{tikzpicture}
	\end{minipage}
	\hspace{0.2cm}
	\begin{minipage}[t]{0.4\linewidth}
		\centering
		\begin{tikzpicture}[scale=0.5]
			\def\myalgslist{\idtcgl, \dirmin, \directt, \direct, \nmso, \birmin, \directc, \xdtcgl$_Q$, \xdtcgl$_L$}
			\pgfplotstableread{frt2.txt}\datatable
			\foreach \j in {0,...,8}{
				\foreach \i in {0,...,8}{
					\pgfplotstablegetelem{\j}{[index]\i}\of{\datatable}
					\let\colorvalue\pgfplotsretval
					\node[fill=blue!\colorvalue!white, draw=black, minimum size=5mm, line width=0.01pt] at (\i,\j) {};
				}
			}
			\foreach \i [count=\xi from 0] in \myalgslist{
				\node[rotate=90, anchor=east] at (\xi,-0.5) {\scriptsize \i};
			}
			\draw[black, very thick, line width=0.5mm] (-0.6,-0.6) rectangle (8.6,8.6);
		\end{tikzpicture}
	\end{minipage}
	\hfill
	\begin{minipage}[t]{0.15\linewidth}
		\centering
		\begin{tikzpicture}[scale=0.5]
			\foreach \v [count=\j from 0] in {0,12.5,25,37.5,50,62.5,75,87.5,100}{
				\node[fill=blue!\v!white,draw=black,minimum size=5mm,line width=0.01pt] at (0,\j) {};
			}
			\foreach \v [count=\j from 0] in {0,15,30,45,60,75,90,105,120}{
				\node[anchor=west] at (0.5,\j) {\v};
			}
			\draw[white, very thick, line width=1mm] (-0.6, -3.75) rectangle (2.5,10);
			\draw[black, very thick, line width=0.5mm] (-0.6,-0.6) rectangle (0.6,8.6);
			
			\node[rotate=90] at (-1.5,4) {Color-bar};
		\end{tikzpicture}
	\end{minipage}
	\caption{Pairwise comparison of algorithms. Each heatmap entry (row, column) indicates how many times the row algorithm outperformed the column algorithm (darker shades indicate more wins). The left panel compares the number of solved problems, while the right panel compares error values, with errors below $10^{-4}$ truncated to $10^{-4}$.}
	\label{fig_FTheatmapa}
\end{figure}
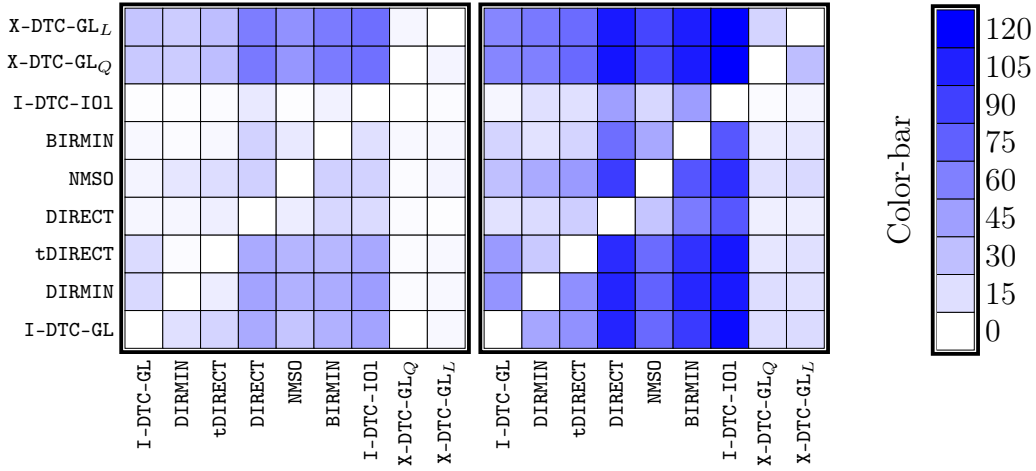

While data profiles provide an aggregated view of the relative performance of the compared algorithms, they do not reveal detailed pairwise relationships between methods on individual problems. 
To complement these analyses, we present pairwise dominance heatmaps in \Cref{fig_FTheatmapa} that summarize how often one algorithm outperforms another across the combined benchmark sets.
The left panel counts problems solved by the row algorithm but not by the column algorithm. 
In contrast, on the left side, it shows how many times the algorithm in the row strictly outperformed the algorithm in the column on the objective error (smaller than $10^{-4}$ are truncated to $10^{-4}$ to avoid overemphasizing differences below the target accuracy threshold). 
Darker shades correspond to higher win counts. 

The left panel shows that the most successful algorithm, \xdtcgl$_Q$, failed on only a very few problems that were solved by other algorithms (\birmin{} and \xdtcgl$_L$ each solved $3$ and $4$ instances, respectively, whereas \xdtcgl$_Q$ failed). 
In contrast, \xdtcgl$_Q$ solved substantially more problems than these methods, namely $29$ and $5$ instances, respectively.

The right panel further confirms the overall advantage of \xdtcgl$_Q$ when considering all problems, including those not solved to the target accuracy. 
Although \xdtcgl$_Q$ is outperformed on $19$, $15$, and $15$ instances by \xdtcgl$_L$, \idtcgl, and \dirmin, respectively, the opposite comparison shows a clear advantage for \xdtcgl$_Q$, with $29$, $54$, and $57$ wins over these methods.

\subsection{Evaluation of Solution Quality Metrics}

\begin{figure}[ht!]
	\centering
	\includegraphics[width=1\linewidth]{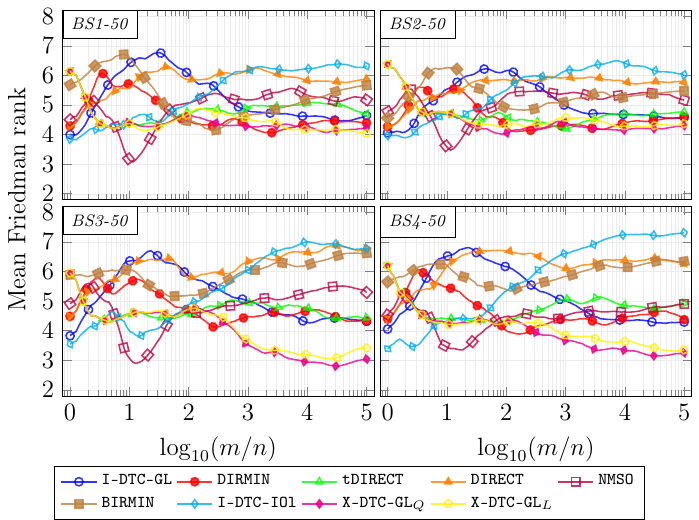}
	\caption{Mean Friedman ranks (lower is better) derived from varying evaluation budgets across four benchmark suites.}
	\label{figs_FrmTest}
\end{figure}

Since some benchmark suites have many unsolved instances, we utilized the Friedman test~\cite{Friedman1937} to rank the algorithms according to the obtained solutions.
The Friedman test was utilized for the varying number of function evaluations, providing insights into the performance of algorithms across different budgets. 
The results of the tests are presented in \Cref{figs_FrmTest}
In this context, a lower mean rank indicates a better algorithm performance.

As the algorithms were generally more successful in addressing the first two benchmark suites, the Friedman mean rank values were more closely aligned than in the last two. 
This is because the algorithms achieved equal ranking on a significant portion of instances in the benchmark suites, thereby reducing the variability in mean ranks.
However, with the maximum evaluation budget, the final algorithm ranking is almost identical, with some exceptions between algorithms in second to fourth places.

For a limited budget (${\leq} n {\times} 10^2$), substantial changes in algorithm rankings are observed, particularly affecting the performance of pure \direct-type algorithms, which tend to exhibit lower rankings. 
Notably, the \nmso{} algorithm performs exceptionally within approximately $n {\times} 10$ function evaluation budgets. 
However, its rankings are substantially worsening as the budget increases. 
\nmso{} consistently secures the fifth position when the evaluation budget reaches the maximum. 
Conversely, the \idtcgl{} algorithm exhibits an inverse pattern, performing poorly within a small evaluation budget but experiencing a significant improvement in ranking as the budget increases. 
Across various benchmark suites, \idtcgl{} consistently claims positions ranging from second to fourth best. 
Finally, our proposed \xdtcgl{} consistently achieves the highest ranking regardless of the surrogate model used, starting from the ${\sim} n {\times} 50$ evaluation budget.

\subsection{Analysis of Execution Time Metrics}

\begin{figure}[ht!]
	\centering
	\includegraphics[width=1\linewidth]{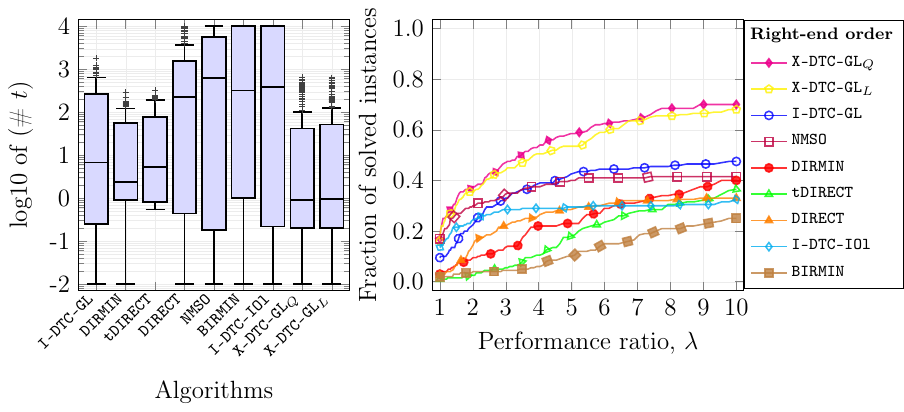}
	\caption{Execution time $t$ comparison of the algorithms on a combined set of $200$ test functions. Execution times $t$ are clipped to $[10^{-2},10^{4}]$ for visualization. The left panel shows boxplots across all instances (lower is better), while the right panel shows a performance profile of solved instances (higher is better).}
	\label{figs_time2}
\end{figure}

To assess the computational efficiency of the considered algorithms, we analyze their execution times ($t$) across a combined benchmark comprising $200$ test functions from four instance sets: \textit{BS1-50}, \textit{BS2-50}, \textit{BS3-50}, and \textit{BS4-50}.
\Cref{figs_time2} summarizes the results using two complementary visualizations.
The left panel shows boxplots of execution times ($t$) across all instances, while the right panel presents a performance profile based on $t$ for successfully solved instances only.
For visualization purposes, execution times are clipped to $[10^{-2},10^{4}]$.

The boxplots show that both the minimum and the first quartile values are below $1$ seconds for all algorithms, indicating that at least $25\%$ of the test functions can be solved at comparable speeds across all solvers.
Furthermore, the median execution times of the two proposed algorithms (\xdtcgl$_Q$, \xdtcgl$_L$) are below one second, meaning that at least $50\%$ of instances are solved within this time.
The median execution times of three other algorithms, \idtcgl, \dirmin, and \directt, are within $10$ seconds ($7.44$, $2.39$, and $5.40$ seconds, respectively), whereas the remaining four algorithms exhibit median values exceeding $200$ seconds.
The third quartile, maximum values, and upper outliers largely correspond to execution times associated with unsolved instances, where algorithms exhausted the entire evaluation budget.
In this context, four algorithms (\dirmin, \directt, and the two proposed algorithms) exhibit similar performance, while the original baseline algorithm, \idtcgl, appears to be slightly slower.
For four algorithms (\direct, \directc, \birmin, and \nmso), execution times $t$ may reach $10^{4}$.
A possible explanation is their relatively low number of evaluations per iteration, which may lead to an excessive number of iterations. 
Because each iteration involves several decision-making computations, the algorithm may require substantial time to exhaust the evaluation budget if a solution is not found early.

To assess how quickly the algorithms identify the global optimum relative to one another, execution time was used to generate performance profiles.
According to the performance profiles (right panel of \Cref{figs_time2}), \nmso{} and \xdtcgl$_Q$ achieve the highest number of wins (i.e., when $\lambda = 1$), solving approximately $17\%$ of the problems in the shortest time.
Another proposed algorithm (\xdtcgl$_L$) also demonstrated strong performance, solving about $16\%$ of the problems the fastest. 
However, as the performance ratio increases to $\lambda{=}2$, the proposed algorithms achieve the highest performance levels among all methods considered. 
Overall, \xdtcgl$_Q$ and \xdtcgl$_L$ outperform the other algorithms on at least $60\%$ of the test problems within a performance ratio of six relative to the best-performing algorithm.

\section{Conclusions and Future Work}
\label{sec_conclusions}

This study introduced a novel approach for solving box-constrained BBO problems by enhancing the \idtcgl{} algorithm with dynamic partitioning and hybridization mechanisms. 
The proposed \xdtcgl{} algorithm constructs local surrogate models within subdivided hyper-rectangles and allocates additional search effort to more promising regions, thereby improving both convergence behavior and solution quality. 
Comprehensive experiments on four benchmark suites based on different ISMs demonstrate that the proposed algorithmic variants outperform relevant alternatives in terms of solution quality, evaluation efficiency, and execution time.
By establishing a substantial performance advancement over the most efficient existing \direct-type baselines, these results implicitly amplify the overall competitiveness of partition-based deterministic optimization within the broader domain of state-of-the-art black-box metaheuristics.

The present study focused on exploiting local monotonicity trends of the objective function and thus partially addressed the concerns raised in~\cite{Jones2021_review} regarding the limited use of local trend information in \direct. 
Future research may investigate more information-driven refinement strategies that adapt the subdivision location according to trends observed in function evaluations. 
Rather than restricting refinement to the geometric center of a selected hyper-rectangle, the search could be directed toward more promising regions, including neighboring hyper-rectangles when surrogate-based indications suggest improvement beyond the current boundaries.

Another promising direction is the investigation of alternative partitioning strategies, such as dividing the longest side into more than three segments (e.g., five, seven, or an adaptive number), which could provide richer sampling for surrogate construction. 
This, in turn, may enable the use of more flexible surrogate models based on the additional sample points. 
Additionally, the observed limitations of the algorithm suggest several directions for future research. 
First, incorporating lower-bound control into the dynamic partitioning strategy may offer further improvements, as indicated by the preliminary observations. 
Second, combining \direct{} with alternative local solvers may enhance convergence, particularly for non-smooth or plateau-rich objective functions where different derivative-free local search methods may exhibit complementary strengths. 
An interesting extension would be the development of an adaptive hybrid framework that dynamically selects or switches between local search procedures based on the observed optimization progress and landscape characteristics.

\section*{Funding}

The research of L. Stripinis was co-funded by the European Union (project ``My First Research Team'', No.~10-092-P-0001, action No.~MPK-49).

\section*{Source code and data availability}

All algorithms, along with the employed configurations and scripts necessary to replicate the findings of this study, are available in the following GitHub repository:
\begin{itemize}
	\item \url{https://github.com/blockchain-group/DIRECTGO}
\end{itemize}

The \directgolib{} test problem library is available as an open-source repository on GitHub and is distributed under the MIT license. 
In addition, the repository contains the necessary codes for the selection of various instance sets used in this study. 
The dataset and codes of the ISMs can be accessed through the following channel:
\begin{itemize}
	\item \url{https://github.com/blockchain-group/DIRECTGOLib}
\end{itemize}

\bibliographystyle{elsarticle-num} 
\bibliography{mybib}

\begin{thebibliography}{10}
\expandafter\ifx\csname url\endcsname\relax
  \def\url#1{\texttt{#1}}\fi
\expandafter\ifx\csname urlprefix\endcsname\relax\def\urlprefix{URL }\fi
\expandafter\ifx\csname href\endcsname\relax
  \def\href#1#2{#2} \def\path#1{#1}\fi

\bibitem{Stork2022_taxonomy}
J.~Stork, A.~E. Eiben, T.~Bartz-Beielstein, A new taxonomy of global
  optimization algorithms, Natural Computing 21 (2022) 219--242.
\newblock \href {https://doi.org/10.1007/s11047-020-09820-4}
  {\path{doi:10.1007/s11047-020-09820-4}}.

\bibitem{Piyavskii1967}
S.~A. Piyavskii, An algorithm for finding the absolute minimum of a function,
  Theory of Optimal Solutions 2 (1967) 13--24, in Russian.
\newblock \href {https://doi.org/10.1016/0041-5553(72)90115-2}
  {\path{doi:10.1016/0041-5553(72)90115-2}}.

\bibitem{Shubert1972}
B.~O. Shubert, A sequential method seeking the global maximum of a function,
  SIAM Journal on Numerical Analysis 9 (1972) 379--388.
\newblock \href {https://doi.org/10.1137/0709036} {\path{doi:10.1137/0709036}}.

\bibitem{Paulavicius2009b}
R.~Paulavi{\v{c}}ius, J.~{\v{Z}}ilinskas, {Global optimization using the
  branch-and-bound algorithm with a combination of Lipschitz bounds over
  simplices}, Technological and Economic Development of Economy 15~(2) (2009)
  310--325.
\newblock \href {https://doi.org/10.3846/1392-8619.2009.15.310-325}
  {\path{doi:10.3846/1392-8619.2009.15.310-325}}.

\bibitem{Paulavicius2010:ol}
R.~Paulavi{\v{c}}ius, J.~{\v{Z}}ilinskas, A.~Grothey, {Investigation of
  selection strategies in branch and bound algorithm with simplicial partitions
  and combination of Lipschitz bounds}, Optimization Letters 4~(2) (2010)
  173--183.
\newblock \href {https://doi.org/10.1007/s11590-009-0156-3}
  {\path{doi:10.1007/s11590-009-0156-3}}.

\bibitem{Jones1993_DIRECT}
D.~R. Jones, C.~D. Perttunen, B.~E. Stuckman, Lipschitzian optimization without
  the lipschitz constant, Journal of Optimization Theory and Applications 79
  (1993) 157--181.
\newblock \href {https://doi.org/10.1007/BF00941892}
  {\path{doi:10.1007/BF00941892}}.

\bibitem{Jones2021_review}
D.~R. Jones, J.~R. R.~A. Martins, The {DIRECT} algorithm: 25 years later,
  Journal of Global Optimization 79~(3) (2021) 521--566.
\newblock \href {https://doi.org/10.1007/s10898-020-00952-6}
  {\path{doi:10.1007/s10898-020-00952-6}}.

\bibitem{Stripinis2023_book}
L.~Stripinis, R.~Paulavi\v{c}ius, {Derivative-free DIRECT-type Global
  Optimization}, SpringerBriefs in Optimization, Springer Cham, 2023.
\newblock \href {https://doi.org/10.1007/978-3-031-46537-6}
  {\path{doi:10.1007/978-3-031-46537-6}}.

\bibitem{Fletcher1987_book}
R.~Fletcher, Practical Methods of Optimization, 2nd Edition, John and Sons
  Chichester, 1987.
\newblock \href {https://doi.org/10.1097/00000539-200101000-00069}
  {\path{doi:10.1097/00000539-200101000-00069}}.

\bibitem{JavadEbadi03072023}
M.~J. Ebadi, A.~Fahs, H.~Fahs, R.~Dehghani, Competitive secant (bfgs) methods
  based on modified secant relations for unconstrained optimization,
  Optimization 72~(7) (2023) 1691--1706.
\newblock \href {https://doi.org/10.1080/02331934.2022.2048381}
  {\path{doi:10.1080/02331934.2022.2048381}}.

\bibitem{nelder1965simplex}
J.~A. Nelder, R.~Mead, A simplex method for function minimization, The Computer
  Journal 7~(4) (1965) 308--313.
\newblock \href {https://doi.org/10.1093/comjnl/7.4.308}
  {\path{doi:10.1093/comjnl/7.4.308}}.

\bibitem{Fadavi2026}
N.~Fadavi, H.~Gangammanavar, Active set-based inexact proximal bundle algorithm
  for stochastic quadratic programming, Computational Optimization and
  Applications 93 (2026) 489--521.
\newblock \href {https://doi.org/10.1007/s10589-025-00739-z}
  {\path{doi:10.1007/s10589-025-00739-z}}.

\bibitem{Kirkpatrick1983_sa}
S.~Kirkpatrick, C.~D. Gelatt, M.~P. Vecchi, Optimization by simulated
  annealing, Science 220~(4598) (1983) 671--680.
\newblock \href {https://doi.org/10.1126/science.220.4598.671}
  {\path{doi:10.1126/science.220.4598.671}}.

\bibitem{Holland1975}
J.~Holland, Adaptation in Natural and Artificial Systems, The University of
  Michigan Press, Ann Arbor, 1975.

\bibitem{kennedy1995particle}
J.~Kennedy, R.~Eberhart, Particle swarm optimization, in: Proceedings of
  ICNN'95-international conference on neural networks, Vol.~4, IEEE, 1995, pp.
  1942--1948.

\bibitem{FU2023119573}
W.-Y. Fu, Accelerated high-dimensional global optimization: A particle swarm
  optimizer incorporating homogeneous learning and autophagy mechanisms,
  Information Sciences 648 (2023) 119573.
\newblock \href {https://doi.org/https://doi.org/10.1016/j.ins.2023.119573}
  {\path{doi:https://doi.org/10.1016/j.ins.2023.119573}}.

\bibitem{Jones1998}
D.~R. Jones, M.~Schonlau, W.~J. Welch, {Efficient Global Optimization of
  Expensive Black-Box Functions}, Journal of Global Optimization 13~(4) (1998)
  455--492.
\newblock \href {https://doi.org/10.1023/A:1008306431147}
  {\path{doi:10.1023/A:1008306431147}}.

\bibitem{ZENG20221641}
Y.~Zeng, Y.~Cheng, J.~Liu, An efficient global optimization algorithm for
  expensive constrained black-box problems by reducing candidate infilling
  region, Information Sciences 609 (2022) 1641--1669.
\newblock \href {https://doi.org/https://doi.org/10.1016/j.ins.2022.07.162}
  {\path{doi:https://doi.org/10.1016/j.ins.2022.07.162}}.

\bibitem{BARTZBEIELSTEIN2017154}
T.~Bartz-Beielstein, M.~Zaefferer, Model-based methods for continuous and
  discrete global optimization, Applied Soft Computing 55 (2017) 154--167.
\newblock \href {https://doi.org/https://doi.org/10.1016/j.asoc.2017.01.039}
  {\path{doi:https://doi.org/10.1016/j.asoc.2017.01.039}}.

\bibitem{Neri2012memetic}
F.~Neri, C.~Cotta, Memetic algorithms and memetic computing optimization: A
  literature review, Swarm and Evolutionary Computation 2~(1) (2012) 1--14.
\newblock \href {https://doi.org/10.1016/j.swevo.2011.11.003}
  {\path{doi:10.1016/j.swevo.2011.11.003}}.

\bibitem{PAN2021304}
J.-S. Pan, N.~Liu, S.-C. Chu, T.~Lai, An efficient surrogate-assisted hybrid
  optimization algorithm for expensive optimization problems, Information
  Sciences 561 (2021) 304--325.
\newblock \href {https://doi.org/https://doi.org/10.1016/j.ins.2020.11.056}
  {\path{doi:https://doi.org/10.1016/j.ins.2020.11.056}}.

\bibitem{Kerschke2019_review}
P.~Kerschke, H.~H. Hoos, F.~Neumann, H.~Trautmann, Automated algorithm
  selection: Survey and perspectives, Evolutionary Computation 27~(1) (2019)
  3--45.
\newblock \href {https://doi.org/10.1162/evco\_a\_00242}
  {\path{doi:10.1162/evco\_a\_00242}}.

\bibitem{DONG2018641}
H.~Dong, B.~Song, P.~Wang, Z.~Dong, {Hybrid surrogate-based optimization using
  space reduction (HSOSR) for expensive black-box functions}, Applied Soft
  Computing 64 (2018) 641--655.
\newblock \href {https://doi.org/https://doi.org/10.1016/j.asoc.2017.12.046}
  {\path{doi:https://doi.org/10.1016/j.asoc.2017.12.046}}.

\bibitem{Stripinis2024_review}
L.~Stripinis, R.~Paulavi{\v{c}}ius, {Review and Computational Study on
  Practicality of Derivative-Free DIRECT-Type Methods}, Informatica 36~(1)
  (2025) 141--174.
\newblock \href {https://doi.org/10.15388/24-INFOR548}
  {\path{doi:10.15388/24-INFOR548}}.

\bibitem{Worbs2025}
T.~Worbs, B.~Rumi, K.~H. Madsen, A.~Thielscher, Realistic electric field
  characterization of clinically used deformable large tms coils in a large
  cohort, Brain Stimulation: Basic, Translational, and Clinical Research in
  Neuromodulation 18 (2025) 1174--1183, doi: 10.1016/j.brs.2025.05.136.
\newblock \href {https://doi.org/10.1016/j.brs.2025.05.136}
  {\path{doi:10.1016/j.brs.2025.05.136}}.

\bibitem{Li2016_DIRECTsqp}
L.~Li, X.~M. Chen, L.~Zhang, A global optimization algorithm for trajectory
  data based car-following model calibration, Transportation Research Part C:
  Emerging Technologies 68 (2016) 311--332.
\newblock \href {https://doi.org/10.1016/j.trc.2016.04.011}
  {\path{doi:10.1016/j.trc.2016.04.011}}.

\bibitem{Kanayama2023_tDIRECT}
K.~Kanayama, A.~Seko, K.~Toyoura, Structure search method for atomic clusters
  based on the dividing rectangles algorithm, PHYSICAL REVIEW E 108 (2023)
  035303.
\newblock \href {https://doi.org/10.1103/PhysRevE.108.035303}
  {\path{doi:10.1103/PhysRevE.108.035303}}.

\bibitem{Dapsys_2023}
I.~Dapšys, R.~Čiegis, V.~Starikovičius, Applying artificial neural networks
  to solve the inverse problem of evaluating concentrations in multianalyte
  mixtures from biosensor signals, Nonlinear Analysis: Modelling and Control
  29~(1) (2023) 53–70.
\newblock \href {https://doi.org/10.15388/namc.2024.29.33604}
  {\path{doi:10.15388/namc.2024.29.33604}}.

\bibitem{HVIDSTEN2025108265}
I.~B. Hvidsten, K.~H. Liland, O.~Tomic, J.~M. Marchetti, Modeling of biodiesel
  production using optimization designs from literature: aiming to reduce the
  laboratory workload, Fuel Processing Technology 275 (2025) 108265.
\newblock \href {https://doi.org/https://doi.org/10.1016/j.fuproc.2025.108265}
  {\path{doi:https://doi.org/10.1016/j.fuproc.2025.108265}}.

\bibitem{CHEN2025114514}
Y.~Chen, F.~Yu, Q.~Zhang, M.~Pratama, Energy-efficient adaptive perception for
  autonomous driving via lightweight policy learning and simulation-based
  optimization, Knowledge-Based Systems 330 (2025) 114514.
\newblock \href {https://doi.org/https://doi.org/10.1016/j.knosys.2025.114514}
  {\path{doi:https://doi.org/10.1016/j.knosys.2025.114514}}.

\bibitem{CHEN20251204}
R.~Chen, X.~Tian, H.~Du, W.~Zhang, Z.~Wang, L.~Xia, J.~Han, K.~Wang, Research
  on generation of toolpaths with smooth tool orientation changes for five-axis
  machining of blisk based on the rotary axes kinematic features of machine
  tool, Journal of Manufacturing Processes 152 (2025) 1204--1219.
\newblock \href {https://doi.org/https://doi.org/10.1016/j.jmapro.2025.08.065}
  {\path{doi:https://doi.org/10.1016/j.jmapro.2025.08.065}}.

\bibitem{Stripinis2024_benchmark}
L.~Stripinis, J.~K\r{u}dela, R.~Paulavi{\v{c}}ius, Benchmarking derivative-free
  global optimization algorithms under limited dimensions and large evaluation
  budgets, IEEE Transactions on Evolutionary Computation 29~(1) (2025)
  187--204.
\newblock \href {https://doi.org/10.1109/TEVC.2024.3379756}
  {\path{doi:10.1109/TEVC.2024.3379756}}.

\bibitem{Jakub2023}
J.~K\r{u}dela, {Benchmarking State-of-the-art DIRECT-type Methods on the BBOB
  Noiseless Testbed}, in: Proceedings of the Companion Conference on Genetic
  and Evolutionary Computation, GECCO'23 Companion, Association for Computing
  Machinery, New York, NY, USA, 2023, pp. 1620--1627.
\newblock \href {https://doi.org/10.1145/3583133.3596308}
  {\path{doi:10.1145/3583133.3596308}}.

\bibitem{bujok2022eigen}
P.~Bujok, P.~Kolenovsky, Eigen crossover in cooperative model of evolutionary
  algorithms applied to cec 2022 single objective numerical optimisation, in:
  2022 IEEE Congress on Evolutionary Computation (CEC), IEEE, 2022, pp. 1--8.

\bibitem{kumar2017improving}
A.~Kumar, R.~K. Misra, D.~Singh, Improving the local search capability of
  effective butterfly optimizer using covariance matrix adapted retreat phase,
  in: 2017 IEEE congress on evolutionary computation (CEC), IEEE, 2017, pp.
  1835--1842.

\bibitem{zhang2018hybrid}
G.~Zhang, Y.~Shi, Hybrid sampling evolution strategy for solving single
  objective bound constrained problems, in: 2018 IEEE Congress on Evolutionary
  Computation (CEC), IEEE, 2018, pp. 1--7.

\bibitem{tanabe2014improving}
R.~Tanabe, A.~S. Fukunaga, Improving the search performance of shade using
  linear population size reduction, in: 2014 IEEE congress on evolutionary
  computation (CEC), IEEE, 2014, pp. 1658--1665.

\bibitem{hadi2021single}
A.~A. Hadi, A.~W. Mohamed, K.~M. Jambi, Single-objective real-parameter
  optimization: Enhanced lshade-spacma algorithm, Heuristics for optimization
  and learning (2021) 103--121.

\bibitem{Stripinis2021_pDIRECT_GLce}
L.~Stripinis, J.~{\v{Z}}ilinskas, L.~G. Casado, R.~Paulavi{\v{c}}ius, On
  {MATLAB} experience in accelerating {DIRECT-GLce} algorithm for constrained
  global optimization through dynamic data structures and parallelization,
  Applied Mathematics and Computation 390 (2021) 125596.
\newblock \href {https://doi.org/10.1016/j.amc.2020.125596}
  {\path{doi:10.1016/j.amc.2020.125596}}.

\bibitem{Tavassoli2015_hddirect}
A.~Tavassoli, K.~H. Hajikolaei, S.~Sadeqi, G.~G. Wang, E.~Kjeang, {Modification
  of DIRECT for high-dimensional design problems}, Engineering Optimization
  46~(6) (2014) 810--823.
\newblock \href {https://doi.org/10.1080/0305215X.2013.800057}
  {\path{doi:10.1080/0305215X.2013.800057}}.

\bibitem{Mockus2017_Plor}
J.~Mockus, R.~Paulavi{\v{c}}ius, D.~Rusakevi{\v{c}}ius, D.~{\v{S}}e{\v{s}}ok,
  J.~{\v{Z}}ilinskas, {Application of Reduced-set Pareto-Lipschitzian
  Optimization to truss optimization}, Journal of Global Optimization 67~(1-2)
  (2017) 425--450.
\newblock \href {https://doi.org/10.1007/s10898-015-0364-6}
  {\path{doi:10.1007/s10898-015-0364-6}}.

\bibitem{Gablonsky2001_DIRECTl}
J.~M. Gablonsky, C.~T. Kelley, A locally-biased form of the {DIRECT} algorithm,
  Journal of Global Optimization 21~(1) (2001) 27--37.
\newblock \href {https://doi.org/10.1023/A:1017930332101}
  {\path{doi:10.1023/A:1017930332101}}.

\bibitem{Stripinis2018_DIRECT_GL}
L.~Stripinis, R.~Paulavi{\v{c}}ius, J.~{\v{Z}}ilinskas, Improved scheme for
  selection of potentially optimal hyper-rectangles in {DIRECT}, Optimization
  Letters 12 (2018) 1699--1712.
\newblock \href {https://doi.org/10.1007/s11590-017-1228-4}
  {\path{doi:10.1007/s11590-017-1228-4}}.

\bibitem{Tao2017_ABC_DIRECT}
Q.~Tao, X.~Huang, S.~Wang, L.~Li, Adaptive block coordinate {DIRECT} algorithm,
  Journal of Global Optimization 69 (2017) 797--822.
\newblock \href {https://doi.org/10.1007/s10898-017-0541-x}
  {\path{doi:10.1007/s10898-017-0541-x}}.

\bibitem{Jones2001_DIRECT_REV}
D.~R. Jones, The {DIRECT} global optimization algorithm, in: C.~A. Floudas,
  P.~M. Pardalos (Eds.), The Encyclopedia of Optimization, Kluwer Academic
  Publishers, Dordrect, 2001, pp. 431--440.

\bibitem{Liuzzi2010_DIRMIN}
G.~Liuzzi, S.~Lucidi, V.~Piccialli, A {DIRECT}-based approach exploiting local
  minimizations for the solution for large-scale global optimization problems,
  Computational Optimization and Applications 45~(2) (2010) 353--375.
\newblock \href {https://doi.org/10.1007/s10589-008-9217-2}
  {\path{doi:10.1007/s10589-008-9217-2}}.

\bibitem{Paulavicius2020_BIRMIN}
R.~Paulavi{\v{c}}ius, Y.~D. Sergeyev, D.~E. Kvasov, J.~{\v{Z}}ilinskas,
  Globally-biased {BIRECT} algorithm with local accelerators for expensive
  global optimization, Expert Systems with Applications 144 (2020) 113052.
\newblock \href {https://doi.org/10.1016/j.eswa.2019.113052}
  {\path{doi:10.1016/j.eswa.2019.113052}}.

\bibitem{Stripinis2019_DIRECT_GLce}
L.~Stripinis, R.~Paulavi{\v{c}}ius, J.~{\v{Z}}ilinskas, Penalty functions and
  two-step selection procedure based {DIRECT}-type algorithm for constrained
  global optimization, Structural and Multidisciplinary Optimization 59 (2019)
  2155--2175.
\newblock \href {https://doi.org/10.1007/s00158-018-2181-2}
  {\path{doi:10.1007/s00158-018-2181-2}}.

\bibitem{Finkel2004_DIRECT_restart}
D.~Finkel, C.~T. Kelley, An adaptive restart implementation of {DIRECT}, Tech.
  Rep. CRSC-TR04-30, North Carolina State University. Center for Research in
  Scientific Computation, online; accessed: 2023-11-08 (2004).

\bibitem{Liu2015_MrDIRECT}
Q.~Liu, J.~Zeng, G.~Yang, {MrDIRECT}: a multilevel robust {DIRECT} algorithm
  for global optimization problems, Journal of Global Optimization 62 (2015)
  205--227.
\newblock \href {https://doi.org/10.1007/s10898-014-0241-8}
  {\path{doi:10.1007/s10898-014-0241-8}}.

\bibitem{Liu2017_MrDIRECT}
Q.~Liu, G.~Yang, Z.~Zhang, J.~Zeng, Improving the convergence rate of the
  direct global optimization algorithm, Journal of Global Optimization 67
  (2017) 851--872.
\newblock \href {https://doi.org/10.1007/s10898-016-0447-z}
  {\path{doi:10.1007/s10898-016-0447-z}}.

\bibitem{Finkel2006_DIRECT_m}
D.~E. Finkel, C.~T. Kelley, Additive scaling and the {DIRECT} algorithm,
  Journal of Global Optimization 36~(4) (2006) 597--608.
\newblock \href {https://doi.org/10.1007/s10898-006-9029-9}
  {\path{doi:10.1007/s10898-006-9029-9}}.

\bibitem{Liu2013_DIRECT_a}
Q.~Liu, Linear scaling and the direct algorithm, Journal of Global Optimization
  56 (2013) 1233--1245.
\newblock \href {https://doi.org/10.1007/s10898-012-9952-x}
  {\path{doi:10.1007/s10898-012-9952-x}}.

\bibitem{Baker2000_Aggressive_direct}
C.~A. Baker, L.~T. Watson, B.~Grossman, W.~H. Mason, R.~T. Haftka, Parallel
  global aircraft configuration design space exploration, in: A.~Tentner (Ed.),
  High Performance Computing Symposium 2000, Soc. for Computer Simulation
  Internat, 2000, pp. 54--66.

\bibitem{Mockus2011_Plo}
J.~Mockus, On the {P}areto optimality in the context of {L}ipschitzian
  optimization, Informatica 22~(4) (2011) 521--536.
\newblock \href {https://doi.org/10.15388/Informatica.2011.340}
  {\path{doi:10.15388/Informatica.2011.340}}.

\bibitem{Stripinis2024_I_DTC_GL}
L.~Stripinis, R.~Paulavi{\v{c}}ius, An empirical study of various candidate
  selection and partitioning techniques in the {DIRECT} framework, Journal of
  Global Optimization 88 (2024) 723--753.
\newblock \href {https://doi.org/10.1007/s10898-022-01185-5}
  {\path{doi:10.1007/s10898-022-01185-5}}.

\bibitem{Stripinis2024_HALRECT}
L.~Stripinis, R.~Paulavi{\v{c}}ius, Lipschitz-inspired {HALRECT} algorithm for
  derivative-free global optimization, Journal of Global Optimization 88 (2024)
  139--169.
\newblock \href {https://doi.org/10.1007/s10898-023-01296-7}
  {\path{doi:10.1007/s10898-023-01296-7}}.

\bibitem{Paulavicius2018_BIRECT}
R.~Paulavi{\v{c}}ius, L.~Chiter, J.~{\v{Z}}ilinskas, Global optimization based
  on bisection of rectangles, function values at diagonals, and a set of
  lipschitz constants, Journal of Global Optimization 71 (2018) 5--20.
\newblock \href {https://doi.org/10.1007/s10898-016-0485-6}
  {\path{doi:10.1007/s10898-016-0485-6}}.

\bibitem{Sergeyev2006_ADC}
Y.~D. Sergeyev, D.~E. Kvasov, Global search based on efficient diagonal
  partitions and a set of lipschitz constants, {SIAM} Journal on Optimization
  16 (2006) 910--937.
\newblock \href {https://doi.org/10.1137/040621132}
  {\path{doi:10.1137/040621132}}.

\bibitem{Guessoum2023_BIRECT_V}
N.~Guessoum, L.~Chiter, Diagonal partitioning strategy using bisection of
  rectangles and a novel sampling scheme, MENDEL 29 (12 2023).
\newblock \href {https://doi.org/10.13164/mendel.2023.2.131}
  {\path{doi:10.13164/mendel.2023.2.131}}.

\bibitem{Paulavicius2013_DISIMPL}
R.~Paulavi{\v{c}}ius, J.~{\v{Z}}ilinskas, Simplicial lipschitz optimization
  without the lipschitz constant, Journal of Global Optimization 59 (2014)
  23--40.
\newblock \href {https://doi.org/10.1007/s10898-013-0089-3}
  {\path{doi:10.1007/s10898-013-0089-3}}.

\bibitem{Holmstrom2004}
K.~Holmstr{\"o}m, M.~M. Edvall, The TOMLAB Optimization Environment, Springer
  US, Boston, MA, 2004, pp. 369--376.
\newblock \href {https://doi.org/10.1007/978-1-4613-0215-5\_19}
  {\path{doi:10.1007/978-1-4613-0215-5\_19}}.

\bibitem{Stripinis2022_DIRECTGO}
L.~Stripinis, R.~Paulavi{\v{c}}ius, {DIRECTGO}: A new {DIRECT}-type {MATLAB}
  toolbox for derivative-free global optimization, {ACM} Transactions on
  Mathematical Software 48 (dec 2022).
\newblock \href {https://doi.org/10.1145/3559755} {\path{doi:10.1145/3559755}}.

\bibitem{DIRECTGOLib2023}
L.~Stripinis, J.~K\r{u}dela, R.~Paulavi\v{c}ius,
  \href{https://github.com/blockchain-group/DIRECTGOLib}{{DIRECTGOLib} -
  {DIRECT} global optimization test problems library}, pre-release v2.0 (2023).
\newline\urlprefix\url{https://github.com/blockchain-group/DIRECTGOLib}

\bibitem{Stripinis2021_DIRECT_GLh}
L.~Stripinis, R.~Paulavi{\v{c}}ius, A new {DIRECT-GLh} algorithm for global
  optimization with hidden constraints, Optimization Letters 15 (2021)
  1865--1884.
\newblock \href {https://doi.org/10.1007/s11590-021-01726-z}
  {\path{doi:10.1007/s11590-021-01726-z}}.

\bibitem{BBOB_Hansen2009}
N.~Hansen, S.~Finck, R.~Ros, A.~Auger,
  \href{https://inria.hal.science/inria-00362633}{Real-parameter black-box
  optimization benchmarking 2009: Noiseless functions definitions}, Research
  Report RR-6829, {INRIA} (2009).
\newline\urlprefix\url{https://inria.hal.science/inria-00362633}

\bibitem{IOHprofiler}
C.~Doerr, H.~Wang, F.~Ye, S.~van Rijn, T.~B{\"a}ck,
  \href{https://arxiv.org/abs/1810.05281}{Iohprofiler: A benchmarking and
  profiling tool for iterative optimization heuristics}, arXiv
  e-prints:1810.05281 (oct 2018).
\newblock \href {http://arxiv.org/abs/1810.05281} {\path{arXiv:1810.05281}}.
\newline\urlprefix\url{https://arxiv.org/abs/1810.05281}

\bibitem{hansen2021coco}
N.~Hansen, A.~Auger, R.~Ros, O.~Mersmann, T.~Tu{\v{s}}ar, D.~Brockhoff, Coco: A
  platform for comparing continuous optimizers in a black-box setting,
  Optimization Methods and Software 36~(1) (2021) 114--144.

\bibitem{conn2009derivativefree}
A.~R. Conn, K.~Scheinberg, L.~N. Vicente, Introduction to Derivative-Free
  Optimization, SIAM, Philadelphia, PA, 2009.
\newblock \href {https://doi.org/10.1137/1.9780898718768}
  {\path{doi:10.1137/1.9780898718768}}.

\bibitem{Abdullah2016_NMSO}
A.~Al-Dujaili, S.~Suresh, A naive multi-scale search algorithm for global
  optimization problems, Information Sciences 372 (2016) 294--312.
\newblock \href {https://doi.org/https://doi.org/10.1016/j.ins.2016.07.054}
  {\path{doi:https://doi.org/10.1016/j.ins.2016.07.054}}.

\bibitem{MATLAB2023}
T.~M. Inc., \href{https://www.mathworks.com}{Matlab version: 9.14.0 (r2023a)}
  (2023).
\newline\urlprefix\url{https://www.mathworks.com}

\bibitem{ABS_Kudela2022}
J.~K\r{u}dela, R.~Matousek, New benchmark functions for single-objective
  optimization based on a zigzag pattern, IEEE Access 10 (2022) 8262--8278.
\newblock \href {https://doi.org/10.1109/ACCESS.2022.3144067}
  {\path{doi:10.1109/ACCESS.2022.3144067}}.

\bibitem{Layeb2022}
L.~Abdesslem, \href{https://www.mathworks.com/matlabcentral}{New hard benchmark
  functions for global optimization}, mATLAB Central File Exchange. Retrieved
  February 18, 2022. (2022).
\newline\urlprefix\url{https://www.mathworks.com/matlabcentral}

\bibitem{liang2013problem}
J.~J. Liang, B.~Y. Qu, P.~N. Suganthan, Problem definitions and evaluation
  criteria for the cec 2014 special session and competition on single objective
  real-parameter numerical optimization, Computational Intelligence Laboratory,
  Zhengzhou University, Zhengzhou China and Technical Report, Nanyang
  Technological University, Singapore 635~(2) (2013).

\bibitem{wu2017problem}
G.~Wu, R.~Mallipeddi, P.~N. Suganthan, Problem definitions and evaluation
  criteria for the cec 2017 competition on constrained real-parameter
  optimization, National University of Defense Technology, Changsha, Hunan, PR
  China and Kyungpook National University, Daegu, South Korea and Nanyang
  Technological University, Singapore, Technical Report (2017).

\bibitem{Stripinis2024ISM}
L.~Stripinis, J.~Kůdela, R.~Paulavičius, Two novel instance selection methods
  combining algorithm performance and landscape analysis: A comparative study
  in continuous optimization, IEEE Transactions on Cybernetics 56~(3) (2026)
  1202--1215.
\newblock \href {https://doi.org/10.1109/TCYB.2025.3625095}
  {\path{doi:10.1109/TCYB.2025.3625095}}.

\bibitem{Cenikj2023}
G.~Cenikj, G.~Petelin, C.~Doerr, P.~Koro\v{s}ec, T.~Eftimov, Dynamorep:
  Trajectory-based population dynamics for classification of black-box
  optimization problems, in: Proceedings of the Genetic and Evolutionary
  Computation Conference, GECCO '23, Association for Computing Machinery, New
  York, NY, USA, 2023, pp. 813--821.
\newblock \href {https://doi.org/10.1145/3583131.3590401}
  {\path{doi:10.1145/3583131.3590401}}.

\bibitem{Dietrich2024Impact}
K.~Dietrich, D.~Vermetten, C.~Doerr, P.~Kerschke, Impact of training instance
  selection on automated algorithm selection models for numerical black-box
  optimization (2024).
\newblock \href {http://arxiv.org/abs/2404.07539} {\path{arXiv:2404.07539}}.

\bibitem{Mersmann2011ELA}
O.~Mersmann, B.~Bischl, H.~Trautmann, M.~Preuss, C.~Weihs, G.~Rudolph,
  Exploratory landscape analysis, in: Proceedings of the 13th Annual Conference
  on Genetic and Evolutionary Computation, GECCO '11, Association for Computing
  Machinery, New York, NY, USA, 2011, pp. 829--836.
\newblock \href {https://doi.org/10.1145/2001576.2001690}
  {\path{doi:10.1145/2001576.2001690}}.

\bibitem{Dolan2002_performance_profiles}
E.~D. Dolan, J.~J. More, Benchmarking optimization software with performance
  profiles, Mathematical Programming 91 (2002) 201--213.
\newblock \href {https://doi.org/10.1007/s101070100263}
  {\path{doi:10.1007/s101070100263}}.

\bibitem{More2009_data_profiles}
J.~J. Mor\'{e}, S.~M. Wild, Benchmarking derivative-free optimization
  algorithms, SIAM Journal on Optimization 20~(1) (2009) 172--191.
\newblock \href {https://doi.org/10.1137/080724083}
  {\path{doi:10.1137/080724083}}.

\bibitem{Friedman1937}
M.~Friedman, The use of ranks to avoid the assumption of normality implicit in
  the analysis of variance, Journal of the American Statistical Association
  32~(200) (1937) 675--701.
\newblock \href {https://doi.org/10.1080/01621459.1937.10503522}
  {\path{doi:10.1080/01621459.1937.10503522}}.

\end{thebibliography}

\end{document}